%% file: main.tex
\documentclass[10pt]{article} 
\PassOptionsToPackage{usenames, dvipsnames}{xcolor}
\usepackage[preprint]{tmlr}

\input{math_commands.tex}

\usepackage{hyperref}
\usepackage{url}
\usepackage{mathtools} 
\usepackage{booktabs} 
\usepackage[dvipsnames]{xcolor}
\usepackage{tikz}
\usepackage{graphicx}
\usepackage{amsmath}
\usepackage{amssymb}

\input{math_cmd}
\usepackage{makecell}
\usepackage{array,booktabs}
\usepackage{tabularx}
\usepackage[algo2e]{algorithm2e}
\usepackage{algorithm}
\usepackage{multirow}
\usepackage{wrapfig}
\usepackage{soul}
\DeclareUnicodeCharacter{2212}{-}

\title{Model Recycling Framework for \\ Multi-Source Data-Free Supervised Transfer Learning}

\author{\name Sijia Wang \email sijia.wang@duke.edu \\
      \addr Department of Electrical and Computer Engineering\\
      Duke University 
      \AND
      \name Ricardo Henao \email ricardo.henao@duke.edu\\
      \addr Department of Electrical and Computer Engineering\\
      Duke University 
}

\def\month{MM}  
\def\year{YYYY} 
\def\openreview{\url{https://openreview.net/forum?id=XXXX}} 

\begin{document}

\maketitle

\begin{abstract}
Increasing concerns for data privacy and other difficulties associated with retrieving source data for model training have created the need for {\em source-free transfer learning}, in which one only has access to pre-trained models instead of data from the original source domains.
This setting introduces many challenges, 
as many existing transfer learning methods typically rely on access to source data, which limits their direct applicability to scenarios where source data is unavailable.
Further, practical concerns make it more difficult, for instance efficiently selecting models for transfer without information on source data, and transferring without full access to the source models.
So motivated, we propose a model recycling framework for parameter-efficient training of models that identifies subsets of related source models to reuse in both white-box and black-box settings.
Consequently, our framework makes it possible for Model as a Service (MaaS) providers to build libraries of efficient pre-trained models, thus creating an opportunity for multi-source data-free supervised transfer learning.
\end{abstract}

\section{Introduction}

Many existing methods for transfer learning and domain adaptation heavily rely on access to data from the source domains \citep{pan2010domain, herath2017learning, kulis2011you, Zhuang2019ACS, yao2010boosting, sun2015subspace, jiang2007instance}. This situation can give rise to privacy concerns, as organizations may not want to share sensitive information; for instance, healthcare providers may be reluctant to share patient information and security system maintainers may not want to risk sharing facial recognition data for system performance updates. Additionally, there may be issues with obtaining the source data such as when it is hard to retrieve due to technical difficulties or intellectual property restrictions \citep{li2020model, Chen2021SelfSupervisedNL, Liang2020DoWR, ahmed2021unsupervised}. 

Recent advancements in source-free unsupervised domain adaptation (SFUDA) have presented solutions for a scenario where source data is not accessible \citep{fang2022source}. Purposely, SFUDA utilizes pre-trained source models to improve the generalization of a model on an unlabeled target dataset. Our work is similar to other approaches in the field of SFUDA \citep{li2020model, Chen2021SelfSupervisedNL, Liang2020DoWR, ahmed2021unsupervised}, in that it addresses the practical scenario where source data is not available during training. 
Importantly, a crucial aspect is often overlooked by the majority of SFUDA studies. When it is assumed that source data is not accessible, then it cannot be guaranteed that the available source models have been trained on domains related to the target task. And yet, most of the works only have experimented on classic domain adaptation benchmarks, which are somewhat related by design, {\em e.g.}, \texttt{Digits-Five} \citep{peng2019moment}, \texttt{Office-31} \citep{saenko2010adapting}, and \texttt{Office-Home} \citep{venkateswara2017deep}, {\em i.e,}, domains that share the same labels but are dissimilar in feature (and ambient) space.

\begin{figure}[t]
\centering
\includegraphics[width=0.9\textwidth]{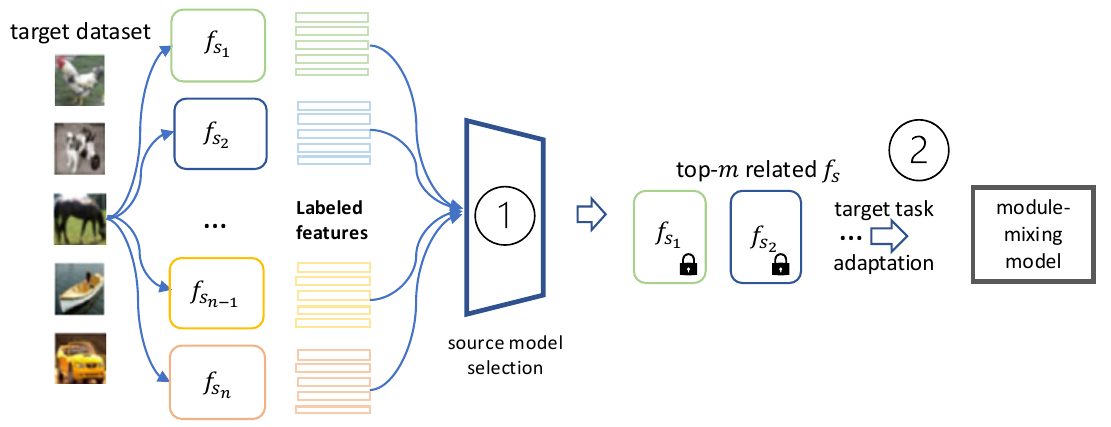}
\caption{Proposed method in a nutshell: \raisebox{0.4mm}{\textcircled{\raisebox{-0.2mm} {\scriptsize 1}}} For the source model selection phase, we extract features for the target dataset with all the source feature extractors $\{\fv_{s_n}\}_{n\in S_m}$, and use these features to select the top-$m$ related source feature extractors. \raisebox{0.4mm}{\textcircled{\raisebox{-0.2mm} {\scriptsize 2}}} For the target task adaptation phase, the selected feature extractors are used to build a module-mixing model in either white-box or black-box settings, the details of which are shown in Figure~\ref{fig:framework}.
} 
\label{fig:pipeline}
\end{figure}

Our approach is unique in that we consider such a {\em source-free supervised transfer learning} (SFSTL) setting \citep{NEURIPS2019_6048ff4e}, where we do not assume source models are trained on tasks with similar feature spaces or labels. Instead, we not only experiment on classic domain adaptation datasets, but also discuss the scenario where source and target tasks have very different datasets or their label sets are partially shared.
Another distinguishing factor of our work is that we consider a framework that can search a pool of pre-trained models for those that are most helpful in enhancing the performance of new tasks.
Furthermore, it also allows one to do transfer with either white-box or black-box assumptions for the source models.
For the {\em white-box scenario}, we assume many pre-trained source models are already available on a server or service, which defines in advance the model architecture to be used by all models. In the {\em black-box scenario}, only the extracted features before the classification head are accessible, as large pre-trained models are considered intellectual properties nowadays, thus model details are sometimes unavailable. In either case, users can tweak parameters, such as the number of source models, learning rate, and weight decay, to obtain a better model relative to training an independent model with their own data. This process would presumably require little technical knowledge. Moreover, since many MaaS platforms would accumulate pre-trained source models over time, we also provide a parameter-efficient solution for training source models to control memory requirements.

Our contributions are summarized below:
{\em i}) We study an under-explored source-free supervised transfer learning scenario, where we are given a collection of related and unrelated source models to improve the performance of a new model on a labeled target task.
Crucially, we only have access to the target data.
{\em ii}) We propose a framework (shown in Figure \ref{fig:pipeline}) that can select several related source models (the number of which is set {\em a priori}), and utilize their knowledge on new tasks in both \emph{white-box} and \emph{black-box} scenarios. We treat these selected models as ‘recycling materials' and reuse them to help learn a better model for the target tasks.
{\em iii}) We perform extensive experiments to highlight the properties of our framework including a specially crafted \texttt{CIFAR100}-based experiment and an ablation study. 

In the next section, we will discuss related works in relation to the proposed work. In Section 3, we define our problem setting and introduce the parameter-efficient finetune method we adopt as the basic building blocks in our model. In Section 4, we explain in detail each component of our model recycling framework, including the training of source models, the identification of most transferable source models, and the module-mixing model design. We also provide the rationale for the design choices of each component in the ablation study and the Appendix. Experiments are presented in Section 5 including the results for the ablation study. For example, we show that the proposed source model selection module is able to select similar tasks to adapt to with good performance in Figure \ref{fig:ctrlknnselection}, and compare our method to other model transferability evaluation methods in Table \ref{table:tinyImagenet}. We also analyze the generalization ability of the proposed model and the visualize the features extracted with the learned model. Finally, conclusions are presented in Section 6.

\section{Related Work}
\paragraph{Multi-source Supervised Transfer Learning}
\citet{tong2021a} presents a transferability measure that is characterized by the sample sizes, model complexity and the similarities between source and target tasks.
Then, they use the transferability measure to form a convex combination of the predictions from different models. Most of the works in this class assume source datasets are available during adaptation \citep{Li2020GradMixMT, Jin2021DeepTL, Xu2018DeepCN, tong2021a, li2022deep}.
Alternatively and most similar to our setting, \citet{NEURIPS2019_6048ff4e, wu2024h} do not assume the availability of source data. Specifically, \citet{NEURIPS2019_6048ff4e} avoid finetuning the source models by leveraging (maximal) correlation analysis and conditional expectation operators to build a classifier from the combined weighting of the feature functions from the source networks.
The inexpensive weighting and lack of finetuning makes it efficient and effective at adapting to multiple source models in the few-shot regime.
Our work is different in that we use a convex combination of feature representations {\em and} task-specific module parameters.
\paragraph{Source-Free Unsupervised Domain Adaptation} \cite{li2020model, Kurmi2021DomainIA, Hou2021VisualizingAK} focus on generating data for source domains to accommodate already existing unsupervised domain adaptation methods. For instance, \cite{Kurmi2021DomainIA} generates proxy source samples by treating the trained source classifier as an energy-based model along with a parametric data generative neural network. \cite{Chen2021SelfSupervisedNL, Liu2021GraphCB, xiong2021source} leverage self-supervised knowledge distillation to finetune source models by adopting a mean-teaching framework \citep{Tarvainen2017MeanTA}. \cite{Liang2020DoWR} use information maximization and self-supervised pseudo-labeling to adapt the target domain to pre-trained source models. Concerning the potential of utilizing diverse knowledge from multiple domains, \cite{liang2021distill, ahmed2021unsupervised, Dong2021ConfidentAnchor, han2023discriminability} explore the possibility of multi-source data-free adaptation. For example, \cite{ahmed2021unsupervised} extends the work from \cite{Liang2020DoWR} by combining the source models with trainable aggregated weights. \cite{Dong2021ConfidentAnchor} introduces a confident-anchor-induced pseudo label generator with multiple source models to provide more reliable pseudo labels for target data. \cite{liang2021distill} studies a challenging scenario where only black-box source models are available during adaptation. Most works in this class assume there is no labeled target data, while we discuss the case when target data is labeled \citep{fang2024source}.
\paragraph{Averaging Model Weights}
Weighted averaging is widely adopted in optimization approaches.
Stochastic Weight Averaging aggregates weights along a single optimization trajectory \citep{Izmailov2018AveragingWL}.
\citet{Matena2021MergingMW} merge models that are fine-tuned on different text classification tasks with the same pre-trained initialization. \citet{Wortsman2022ModelSA} focus on averaging weights across independent runs of models on the same dataset.
In their ``cross-dataset soup'', models are trained on different datasets and adapted to target tasks by learning a set of averaging weights.
Another similar work \citet{Rame2022ModelRR} also reuses multiple pretrained foundation models to adapt to a new task. The difference is that they finetune the pretrained foundation models on the new task before averaging their weights to create the final model, and assumes that all the pretrained models have the same architecture.
Our work is different in that new modules allocated for target tasks are {\em trained together} with the mixing weights to learn task-specific features. 
Further, we also discuss how to mix features from different models when their dimensions are different in a black-box scenario.
\paragraph{Finetuning Models with Task-Specific Modules}
As deep learning models grow in size \citep{He2015DeepRL, Dosovitskiy2020AnII, radford2019language}, storing and finetuning the whole model becomes exceedingly challenging, due to needing expensive computational resources.
Therefore, there is a line of work proposing that instead of finetuning all the parameters in a pre-trained model, one can instead partition the network into a frozen backbone model and task-specific modules for finetuning.
The idea has been applied to generative models for synthesizing images \citep{Perez2018FiLMVR, cong2020gan, Verma2021EfficientFT}, discriminative models for image classification \citep{Verma2021EfficientFT}, and finetuning for downstream tasks with large language models \citep{Houlsby2019ParameterEfficientTL, li2021prefix}.
This concept is used in building our white-box parameter-efficient models.

\paragraph{Model Transferability Evaluation}
There are studies focused on assessing the transferability of trained models for a specific task. For instance, the pioneer work LEEP \citep{nguyen2020leep}, measures the log expectation of the empirical predictor by estimating the joint distribution across pretrained labels and the target labels. It features fast selection speed; however, it requires the availability of the source models’ classification heads. LogME proposes to estimate the maximum value of label evidence given features extracted by trained models and obtained the Logarithm of Maximum Evidence (LogME) measure \citep{you2021logme}. It exceeds LEEP and NCE \citep{tran2019transferability} in terms of evaluation accuracy, and only uses extracted features and labels to assess source models. However, it still needs extra training to determine the best suited model, which makes this approach impractical when the model pool is too large. \cite{guo2023identifying} recognizes a similar challenge as in our work, which is how can a user search for useful models in the {\em learnware} market. They identify useful models simply by comparing user requirements with model specifications, without running models. However, they need the model developers to submit specifications of their model, which involves generating a feature matrix that describes their data with a public feature extractor. In practice, the model specification details might not always be available. Our choice of model selection features a non-parametric $k$-NN method, which offers a trade-off between selection efficiency and accuracy. We note that the model selection module of our framework can be used interchangeably with the aforementioned methods.
Note however that several factors need to be considered when choosing which one to use, including the size of the model pool, the transferability accuracy demand of the users, and also, whether the source model information is limited.

\section{Background}
\subsection{Problem Setting} \label{sec:problem setting}
In this work, we aim to address the challenge of adapting a collection of classification models trained on different source tasks to a new target task. The goal is to optimize the classification accuracy of the target task. Specifically, consider we have $N$ source models $\{\hv_{s_1}, \hv_{s_2}, \cdots, \hv_{s_N}\}$ corresponding to $N$ source tasks $\{\CT_{s_1}, \CT_{s_2}, \cdots, \CT_{s_N}\}$, and
that we also have a target task $\CT_{t}$ with a labeled dataset $\{(\CX_{t}, \CY_{t})\}$.
Importantly, during training for $\CT_t$, we only have access to the target task dataset $\{(\CX_{t}, \CY_{t})\}$ and source models $\{\hv_{s_1}, \hv_{s_2}, \cdots, \hv_{s_N}\}$.
Each source model consists of a feature extractor $\fv$ and the classification head $\cv$. We denote the source model as $\hv_{s_n}:=(\fv_{s_n} \circ \ \cv_{s_n})$, with $\fv_{s_n}: \CX_{t} \rightarrow \mathbb{R}^{d_{s_n}}$ being the feature extractor with output feature vector of dimension $d_{s_n}$, $\cv_{s_n}: \mathbb{R}^{d_{s_n}} \rightarrow \mathbb{R}^{\alpha}$ being a $\alpha$-way classifier and $\circ$ denoting function composition. The goal is to learn a classification model $\hv_t$ with all the available knowledge.
In practice, source and target tasks do not need to be restricted to having the same number of classes since we only use source models as feature extractors.

\subsection{Efficient Feature Transformation for Task-Specific Modules}\label{feature transformation}
In the white-box setting, we propose to build models with task-specific modules for source and target tasks because they are memory efficient and less likely to overfit on small datasets.
Specifically, we adopt a task-specific module design termed efficient feature transformation (EFT) \citep{Verma2021EfficientFT} for our source and target models. 
EFT proposes appending a small convolutional transformation to each convolutional layer's output feature maps.
The transformation involves two kinds of convolutional kernels that capture spatial features within groups of channels ({\em group-wise filter}) and features across channels at every pixel in the feature map ({\em point-wise filter}).
Given the feature maps $F \in \mathbb{R}^{M \times Q \times K}$ of a layer, with $K$ being the number of feature maps, and $M$ and $Q$ being the spatial dimensions of each feature map, to apply a {\em groupwise filter} $\omega^{s}_i \in \mathbb{R}^{3 \times 3 \times a}$,  we need to first split $F$ into $K/a$ groups of $a$ feature maps. We then convolve a unique filter $\omega^s_i$ with each group and get a new group of feature maps $H^s_i \in \mathbb{R}^{M \times Q \times a}$ ({\em i.e.}, each group of feature maps has a unique spatial filter). Subsequently, we concatenate all the $K/a$ new feature maps $H^s_i$ into $H^s \in \mathbb{R}^{M \times Q \times K}$, which has the same dimension as old feature maps $F$. To apply a {\em point-wise filter} $\omega^{d}_{i} \in \mathbb{R}^{1 \times 1 \times b}$, the same procedures as above need to be applied to split $F$ into $K/b$ groups of $b$ feature maps to create $H^d \in \mathbb{R}^{M \times Q \times K}$. The final feature maps are $H = H^s + \gamma H^d$, where $\gamma \in \{ 0, 1\}$ indicates if the point-wise filters are used or not. By setting $a\ll K$ and $b\ll K$, the amount of trainable parameters is substantially reduced. For instance, using a ResNet18 backbone, $a=8$, and $b=16$ results in 449k parameters per new task, which is $3.9\%$ the size of the backbone. Additional details can be found in the Appendix \ref{A.1} and in \citet{Verma2021EfficientFT}.

\begin{figure}[t]
\centering
\includegraphics[width=\textwidth]{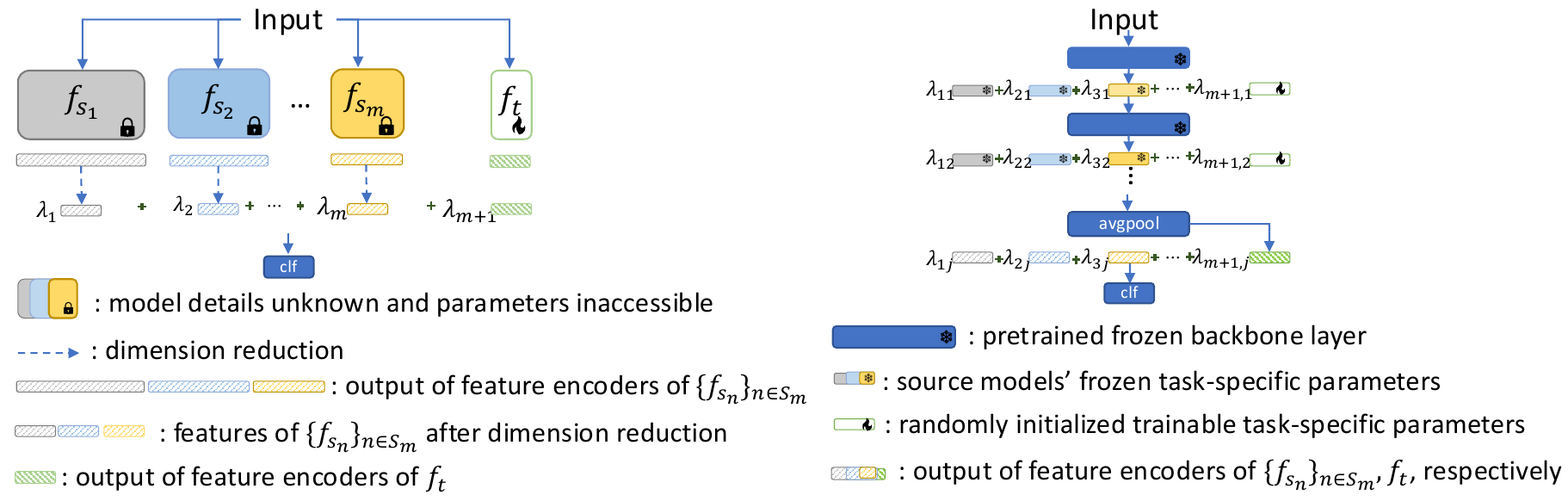} 
\caption{Proposed module-mixing model. Left (black-box): The output features from the source models are first reduced to match the feature dimension of the target model, and then combined with the target features as in \eqref{eq:feat_comb}; Right (white-box): For each task-specific layer, modules from source models (frozen during training), and a randomly initialized EFT module are combined as in \eqref{eq:param_comb}. The output features of the source models and that of the new target model are also aggregated as in \eqref{eq:feat_comb}.
}
\label{fig:framework}
\end{figure}

\section{Proposed Method}
To highlight the flexibility of the framework, we discuss both white-box and black-box assumptions for our source models. We provide a parameter-efficient solution for source model generation. However, we emphasize that the framework is not restricted to using this design, which is why we provide a more flexible setting for the black-box scenario, where we use different pre-trained APIs as our source models. Within both scenarios, the adaptation model in Figure \ref{fig:framework} is obtained through two steps. \emph{i)} Given a large pool of $N$ pre-trained source models, it is not practical nor efficient to learn (transfer) from all of them for a new task $\CT_t$. Therefore, we first aim to select a subset of $m$ models that are likely trained on related source domains, from which $\CT_t$ can learn to improve performance.
\emph{ii)} After the selection process, instead of fine-tuning the pre-trained models directly, we propose a method dubbed {\em module-mixing} for target task adaptation. 

\textbf{White-box scenario}: all models are trained with a modular architecture sharing a common pre-trained backbone. For each task-specific layer, we perform a convex combination on {\em a)} all the task-specific modules from selected models and a randomly initialized task-specific module; {\em b)} the output of the selected source feature extractors and that of the new model.

\textbf{Black-box scenario}: we only have access to the output features from source models, and we are transferring to a target model that may have a different (and potentially smaller) architecture from the source models. Specifically, we first reduce the dimensionality of the source model features to match that of the target model, and then aggregate them using a convex combination.
\subsection{Source Model Generation} \label{sec:source model}
For the white box scenario in which we can control the source model generation, we propose using EFT to build a large library of source models to keep computation and memory requirements under control.
Assume there are labeled datasets $\{(\CX_{s_n}, \CY_{s_n})\}$ for tasks $\CT_{s_n}$, $n \in \{1, \cdots, N\}$, with $\CX_{s_n}$ and $\CY_{s_n}$ being the feature and label domains, respectively.
The feature extractor $\fv_{s_n}$ consists of a backbone $\phiv_{bb}$ that is shared amongst all the source models and a set of task-specific modules $\{\thetav_{s_{n},j}\}^J_{j=1}$, with $j$ being the task-specific layer index and $J$ being the total number of task-specific layers in the model. Below we write $\thetav_{s_n}$ for simplicity. We train model $\hv_{s_n}$ by minimizing a standard cross-entropy loss:
\begin{align}
    \thetav^{*}_{s_n}, \cv^{*}_{s_n} 
    &= \argmin_{\thetav_{s_n}, \cv_{s_n}} \ \CL(\phiv_{bb}, \thetav_{s_n}, \cv_{s_n}; \CX_{s_n}, \CY_{s_n}) \notag \\
    &= - \mathbb{E}_{(x,y)\in \CX_{s_n}\times\CY_{s_n}} \textstyle{\sum}^{\alpha}_{i=1} t_i \log (\delta_i(\hv_{s_n}(x))), \notag
\end{align}

where $\phiv_{bb}$ is frozen during training, $\thetav^{*}_{s_n}$ and $\cv^{*}_{s_n}$ are the optimized task-specific modules and classifier, respectively, $t_i=1$ for the corresponding class $i$ in one-hot encoding {(of $\alpha$ classes), and $\delta_i(\ov) = \frac{exp(\ov_i)}{\sum^{m}_{j=1} exp(\ov_j)}$ is $i$-th element of the softmax activation output vector $\ov=\hv_{s_n}(x)$. We also give further training details, such as the learning rate and batch size in Appendix \ref{A.2} Hyperparameter settings.
\subsection{Selection of Related Source Models}
Given the target training dataset $\{(\CX_{t}, \CY_{t})\}$, we can extract a dataset of features by collecting the output of the feature extractors $\{\fv_{s_n}\}^N_{n=1}$. Each feature dataset is denoted as $D_{t, s_n}:= \{\fv_{s_n}(\CX_{t}), \CY_{t}\}$, for $n \in \{1, \cdots, N\}$.
The goal is to select the top-$m$ corresponding source models (denoted as $S_m$) with the best $k$-NN classifier validation accuracy via data from each $D_{t, s_n}$. The rationale is that a higher validation accuracy indicates that the feature extractor used for extracting features was trained on a domain more related to the target task.
We consider different values of $m$ in the experiments.
Below are the details of our $k$-nearest neighbor ($k$-NN) \citep{knn, 9119820} classifier.
We denote the set of nearest neighbors of $\fv_{s_n}(\xv)$ as $S_{\xv, s_n}$.
In the experiment we set $k=5$. 

For a given validation point $\xv$, we obtain $\fv_{s_n}(\xv_i{''}) \in S_{\xv, s_n}, i \in \{1, \cdots, k\}$ that satisfies
\begin{align}
    {\rm dist}(\fv_{s_n}(\xv), \fv_{s_n}(\xv')) 
    \geq \max \{ {\rm dist}(\fv_{s_n}(\xv), \fv_{s_n}(\xv_i{''})) \}, \ \ \
    \forall(\fv_{s_n}(\xv{'}), \yv{'}) &\in D_{t, s_n} \setminus S_{\xv, s_n} ,
\end{align}
where ${\rm dist}(\cdot,\cdot)$ is a distance metric. We use $D_{t, s_n} \setminus S_{\xv, s_n}$ to represent the subset of $D_{t, s_n}$ excluding  $S_{\xv, s_n}$.
In this work, we employ the Euclidean distance:
$$
    {\rm dist}(\fv_{s_n}(\xv), \fv_{s_n}(\xv{'})) = \left(\sum^d_{i=1} |\fv_{s_n}(\xv)_i - \fv_{s_n}(\xv{'})_i|^2\right)^{\frac{1}{2}}.
$$
Moreover, the $k$-NN classifier $g_{\text{KNN}}$ is defined as 
\begin{align}
    g_{\text{KNN}} = \mathrm{mode}(\yv{''}: (\fv_{s_n}(\xv_i{''}), \yv{''}) \in S_{\xv, s_n}) ,
\end{align}
where $\mathrm{mode}(\cdot)$ takes the label value that occurs most frequently in the set $S_{\xv, s_n}$. $g_{\text{KNN}}$ will predict the label for each validation point $\xv$ based on a majority voting scheme with the nearest neighbors in $S_{\xv, s_n}$. 

\paragraph{Remarks} $k$-NN leverages the majority voting of nearest neighbors when predicting the class labels. A higher $k$-NN score means most of the nearest neighbors have the same label as the sample’s ground truth, which indicates the feature space created by the source feature extractor presents the classes in the target task better. Furthermore, when selecting useful sources, our main concern is to find a solution that is {\em flexible} to both white-box and black-box scenarios, while {\em maintaining its efficiency} when having to deal with selecting from a large pool of models.
$k$-NN, with its non-parametric property, offers insights of how transferable the source models are to the target tasks without any training, which makes it an efficient searching method.

\subsection{Module-Mixing with Distance Correlation Loss}
\paragraph{White-box Scenario} In this setting, the architecture for all models is defined in advance for convenience.
After selecting the top-$m$ source models, we will use them to build the module-mixing model for the target task $t$. As mentioned above, each pre-trained feature extractor $\fv_{s_n}$ has task-specific weights $\{\thetav_{s_n, j}\}^{J}_{j=1}$. For each task-specific layer $j$, the weights for the target task $t$ are
\begin{equation}\label{eq:param_comb}
    \thetav_{t, j} = \lambda_{m+1, j} \thetav_{new} + \sum^{}_{n\in S_m} \lambda_{n,j} \thetav_{s_n, j} ,
\end{equation}
with $\thetav_{new}$ being a randomly initialized new module, $\thetav_{s_n,j}$ the frozen task-specific module of layer $j$ of a selected source model from $S_m$, and $\sum^{m+1}_{n=1} \lambda_{n, j} = 1, \forall j \in \{1, \cdots, J\} $. 
Further, the combined feature representation of the module-mixing model is 
\begin{equation}\label{eq:feat_comb}
     \fv(\cdot) = \lambda_{m+1, J+1} \fv_t(\cdot) + \sum^{}_{n\in S_m} \lambda_{n, J+1}\fv_{s_n}(\cdot) ,
\end{equation}
with $\{\fv_{s_n}(\cdot)\}^{}_{n\in S_m} \in \BR^{\cdot \times d_{s_n}}$ and $\fv_t(\cdot) \in \BR^{\cdot \times d_t}$ being the features (outputs before the classification head) of selected source models and that of the target model, respectively, and $\sum^{m+1}_{n=1} \lambda_{n, J+1} = 1$.
Note that $\{\lambda_{n,j}\}^{m+1}_{n=1}, j \in \{1, \cdots, J+1\}$, denoted as $\lambdav_{t}$ for simplicity, are parameters that can be adjusted at initialization and then learned. By setting the initial values of $\lambdav_{t}$, we can control the importance of each task. For instance, we can assume having no prior information on the task importance by initializing $ \lambdav_{t}$ in all layers with a uniform distribution as done below in the experiments.

\paragraph{Black-box Scenario} We now assume the source models are all black-box APIs, which only produce the features before the classification head. We choose a much smaller architecture for the target model with feature dimensionality being smaller than that extracted from the black-box APIs. The rationale behind this choice is that we want to prevent overfitting provided that the target task usually has a relatively small dataset.
Since the extracted feature dimensions from models can be different, we align the dimensions of the APIs' output features to that of the target model via FastICA \citep{hyvarinen1999fast, hyvarinen2000independent}, an efficient algorithm for independent component analysis. Then we combine the features as in \eqref{eq:feat_comb}. In the experiments, we show that we can still benefit from such a simplified strategy.

\paragraph{Distance Correlation Loss}
We expect the new modules for the target task can learn knowledge that is not in the mixing modules from the source tasks. Specifically, if we can encourage the learned features from the new modules to be more independent from that of the source models, we could potentially achieve better performance. One way to do so is to train the target module-mixing model with distance correlation loss, which was proposed by \citet{Zhen2022OnTV} as a loss function for improving robustness of neural networks against adversarial attacks.

Distance correlation (DC) \citep{Szekely2007MeasuringAT} is a measure of dependence between random vectors. DC between two random variables $X \in \BR^p$ and $Y \in \BR^q$ satisfies $0 \leq DC \leq 1$, and $DC = 0$ if and only if $X$ and $Y$ are independent. Further, $DC(X, Y)$ is defined for $X$ and $Y$ in arbitrary dimensions. This property allows one to minimize DC between $\{\fv_{s_n}\}_{n\in S_m}$ and $\fv_{t}$ even when their output feature dimensions $d_{s_n} \neq d_{t}$.
Here we follow the notation by \citet{Szekely2007MeasuringAT}. We use a stochastic estimate of DC by averaging over minibatches $\xv$ with $n$ samples each. 
The objective for minimizing the distance correlation is 
\begin{align}\label{eq:dc loss}
    \frac{\left<\Av(\fv_{s_n}; \xv), \Bv(\fv_t;\xv) \right>}{\sqrt{\left<\Av(\fv_{s_n}; \xv), \Av(\fv_{s_n}; \xv)\right> \left< \Bv(\fv_t;\xv), \Bv(\fv_t;\xv)\right>}},
\end{align}
where $\left<\Av, \Bv\right>=\sum_{i,j}^{n}(\Av)_{i,j}(\Bv)_{i,j}$, $\Av(\fv_{s_n}; \xv) \in \BR^{n\times n}$ (simplified as $\Av$) and $\Bv(\fv_{t}; \xv) \in \BR^{n\times n}$ (simplified as $\Bv$) are distance matrices computed with $X := \fv_{s_n}(\xv) \in \BR^{n \times d_{s_n}}$ and $Y := \fv_{t}(\xv) \in \BR^{n \times d_{t}}$, respectively, with
\begin{align}
    a_{k,l} = ||X_k - X_l||, \ \ \bar{a}_{k, \cdot}=\frac{1}{n}\sum^{n}_{l=1}a_{k, l}, \ \ \bar{a}_{\cdot, l}=\frac{1}{n}\sum^{n}_{k=1}a_{k, l}, \notag \\
    \bar{a}_{\cdot, \cdot}=\frac{1}{n^2}\sum^{n}_{k,l=1}a_{k, l}, \ \ A_{k,l}=a_{k,l}-\bar{a}_{k,\cdot}-\bar{a}_{\cdot,l}+\bar{a}_{\cdot, \cdot} \notag,
\end{align}
where $k, l \in \{1, \cdots, n\}$, and $A_{k,l}$ is the $k^{th}$ row and $l^{th}$ column of $\Av$. Similarly, we can define $b_{k,l} = ||Y_k-Y_l||$, and $B_{k,l} =
b_{k,l}-\bar{b}_{k,\cdot}-\bar{b}_{\cdot,l}+\bar{b}_{\cdot,\cdot}$.

We optimize all the module-mixing models with the cross-entropy loss and DC loss as below, with $\sigma$ being a constant trade-off parameter:
\begin{equation}\label{l_total}
    \CL_{total} = \CL_{CE} + \sigma \sum_{n \in S_{m}} \CL_{DC} (\fv_{s_n}(\cdot), \fv_t(\cdot)) ,
\end{equation}

The objective $\thetav^{*}_{new}, \cv^{*}_{t}, \lambdav_{t}^* = \argmin \CL_{total}(\hv_{t}; \CX_{t}, \CY_{t})$  indicates that the trainable parameters in $\hv_t$ are the new modules $\thetav_{new}$, the classification head $\cv_{t}$ and convex combination parameters $\lambdav_{t}$. The detailed algorithm is presented in Algorithm \ref{alg:module-mixing}. 
Overall, our motivation for using distance correlation loss is that it encourages features extracted from previous models to be independent from the features learned from the new tasks; Furthermore, since the output feature dimensions for new and old tasks might not be the same, we need the loss to be able to handle features of different dimensions.

The reasons for using this module-mixing technique are threefold. First, it addresses feature saturation and the issue of plasticity loss that occurs when model weights are fine-tuned over time, which is discussed in \cite{dohare2021continual, Ash2019OnTD}. They argue that, in a continual learning setting, 
if a model is finetuned sequentially on several tasks, the model will likely not benefit from random initialization for later tasks as it will lose plasticity over time. Therefore, our goal is to devise a way to make use of pre-trained models rather than attempting to modify them. Our model ensures that we can learn from previous knowledge while also adding new capacity at the start of training for each new task. Secondly, the flexible nature of the model enables us to investigate whether there is a benefit to learning from more tasks, how to balance training and inference efficiency, and the number of tasks to be reused. The reason is that when the number of selected tasks increases, the amount of trainable parameters of the model grows slowly since only $\lambdav_t$ is growing, which is negligible relative to $\thetav_{new}$ and $\cv_{t}$.
Importantly, the last and third reason is that this approach allows us to transfer whether the source models are treated as white- or black-boxes.

\begin{algorithm}[t]
\caption{Module-mixing Model.}\label{alg:module-mixing}
\KwData{Training data $D_{t} = \{\CX_t, \CY_t\}$; Selected source models $\{\fv_{s_n}\}_{n\in S_m}$;}
\KwResult{White-box: Module-mixing model $\fv$ constructed with \eqref{eq:feat_comb} and \eqref{eq:param_comb}; Black-box: Module-mixing model $\fv$ constructed with only \eqref{eq:feat_comb}} 
\BlankLine
\For{$epoch\leftarrow 1$ \KwTo $C$}{
    \For{each minibatch $\xv$}{
    \For {$n \leftarrow 1$ \KwTo $S_m$}{
    Calculate distance matrices $\Av(\fv_{s_n}; \xv)$ and $\Bv(\fv_{t}; \xv)$ in equation \eqref{eq:dc loss}
    
    Calculate $\CL_{DC} (\fv_{s_n}(\xv), \fv_t(\xv))$ as in \eqref{eq:dc loss}
    }
    Calculate $\CL_{total}$ as in \eqref{l_total}
    
    Update ${\thetav_{new}, \cv_{t}, \lambdav_{t}}$ using $\thetav^{*}_{new}, \cv^{*}_{t}, \lambdav_{t}^* = \argmin \CL_{total}(\hv_{t}; \CX_{t}, \CY_{t})$
    }
}
\end{algorithm}
\section{Experiments}
\paragraph{Datasets}
We create three main tasks with \texttt{CIFAR100} \citep{krizhevsky2009learning}, \texttt{Office-31} \citep{saenko2010adapting}, and the $S^{long}$ stream in \texttt{CTrL} (Continual Transfer Learning benchmark) \citep{veniat2020efficient}. 
Each dataset represents a unique scenario. With a special task creation scheme for \texttt{CIFAR100} (detailed below), we study the cases where the classes in each task overlap or are completely different. With \texttt{Office-31}, each task has the same labels but different data distributions.
With the challenging \texttt{CTrL}, we have a large pool of models from which to start, and source and target tasks come from very distinct datasets and have diverse sample sizes. Moreover, an experiment on \texttt{Tiny-ImageNet} \citep {le2015tiny} is shown in the ablation study.

\paragraph{Network Architecture and Implementation Details}
For $k$-NN source model selection, we set $k=5$ for all experiments, then select the source models with the top-$m$ highest $k$-NN scores based on the results of each task's validation set. One experiment discussing other settings for $k$ is provided in the Appendix \ref{A.5}. {\em (i)} In the white-box setting, for source model generation and target task adaptation, we choose a pre-trained ResNet-18 on ImageNet as our frozen backbone and train the EFT modules with the Adam optimizer. We set $a=8$ and $b=16$ for EFT for experiments on \texttt{CIFAR100} and \texttt{Office-31}, and $a=2$ and $b=1$ for EFT on \texttt{CtrL}. {\em (ii)} In the black-box setting, a simple LeNet-5 \citep{lecun1998gradient} is used as target model. We provide results for when the source models are ($a$) ResNet-18s with EFT and ($b$) a pre-trained MAE \citep{He2021MaskedAA} (written as API in the tables). The DC loss trade-off is set to $\sigma=0.05$ in all experiments. 
Implementation details can be found in the Appendix \ref{A.2}.

\paragraph{Baselines} $(a)$ \textbf{Independent Model:} Add randomly initialized EFT modules to a pre-trained frozen backbone and train the model solely with target task data. $(b)$ \textbf{Multi-Source SVM}: Extract features from $m$ randomly selected source models with the target training set, then concatenate the $m$ set of extracted features and use them to train an SVM model \citep{cortes1995support}. $(c)$ \textbf{Maximum Correlation Weighting (MCW)}: Learn the maximum correlation parameters of $m$ randomly selected source feature extractors with one pass of the target training data \citep{NEURIPS2019_6048ff4e}. $(d)$ \textbf{Finetune Source}: Take the source model selected by $k$-NN with $m=1$, and finetune the corresponding task-specific modules. $(e)$ \textbf{Cross-dataset Soup}: Train an independent model for the target task first, and then train the mixing weights with selected source models \citep{Wortsman2022ModelSA} with top-$m$ $k$-NN scores.
$(f)$ \textbf{Model Stacking}: Make predictions with the selected top-$m$ source models and an independent model for the target task, then use those predictions as input features in a logistic regression model.
$(g)$ \textbf{DECISION}: Freezes the last classification layers of source models and jointly optimize the source feature encoders together with their aggregated weights. The source classification layers are used to generate pseudo labels for target data \citep{Ahmed2021UnsupervisedMD}. $(h)$ \textbf{DATE (SHOT)}: Utilizes a source-similarity transferability module and a proxy discriminability perception module for weight determination for source models \citep{han2023discriminability}.
$(i)$ \textbf{DINE}: First distills knowledge from the source predictor to a customized target model, then fine-tune the distilled model to further fit to the target domain \citep{Liang2021DINEDA}.

\textbf{MCW}, \textbf{DECISION}, \textbf{DATE}, \textbf{DINE} are all previous state-of-the-art (SOTA) methods. However, only \textbf{MCW} is designed for the same multi-source data-free supervised transfer learning setting (target task has labels). \textbf{DECISION} and \textbf{DATE} have the most similar problem setting as ours, which is multi-source data-free unsupervised domain adaptation (target task does not have labels). Furthermore, \textbf{DINE} is proposed for adapting to single-source and multi-source black-box models when target task does not have labels. The results of \textbf{DECISION}, \textbf{DATE} and \textbf{DINE} are obtained by strictly conforming to their public source code.

Though \textbf{DECISION}, \textbf{DATE (SHOT)}, \textbf{DINE} 
achieve great performance under each of their settings, they all assume close-set label spaces between source and target domains. Besides, due to their unsupervised nature, they all use pseudo labels predicted by the source models to train the target models. According to \cite{Yi2023WhenSD}, when the noise rate in pseudo labels is high at the beginning of the training, the target model will quickly remember the noise due to confirmation bias. All the above reasons explain the performance deterioration of these methods under our setting.

\begin{figure}[t]
   \centering
   \includegraphics[width=0.45\textwidth]{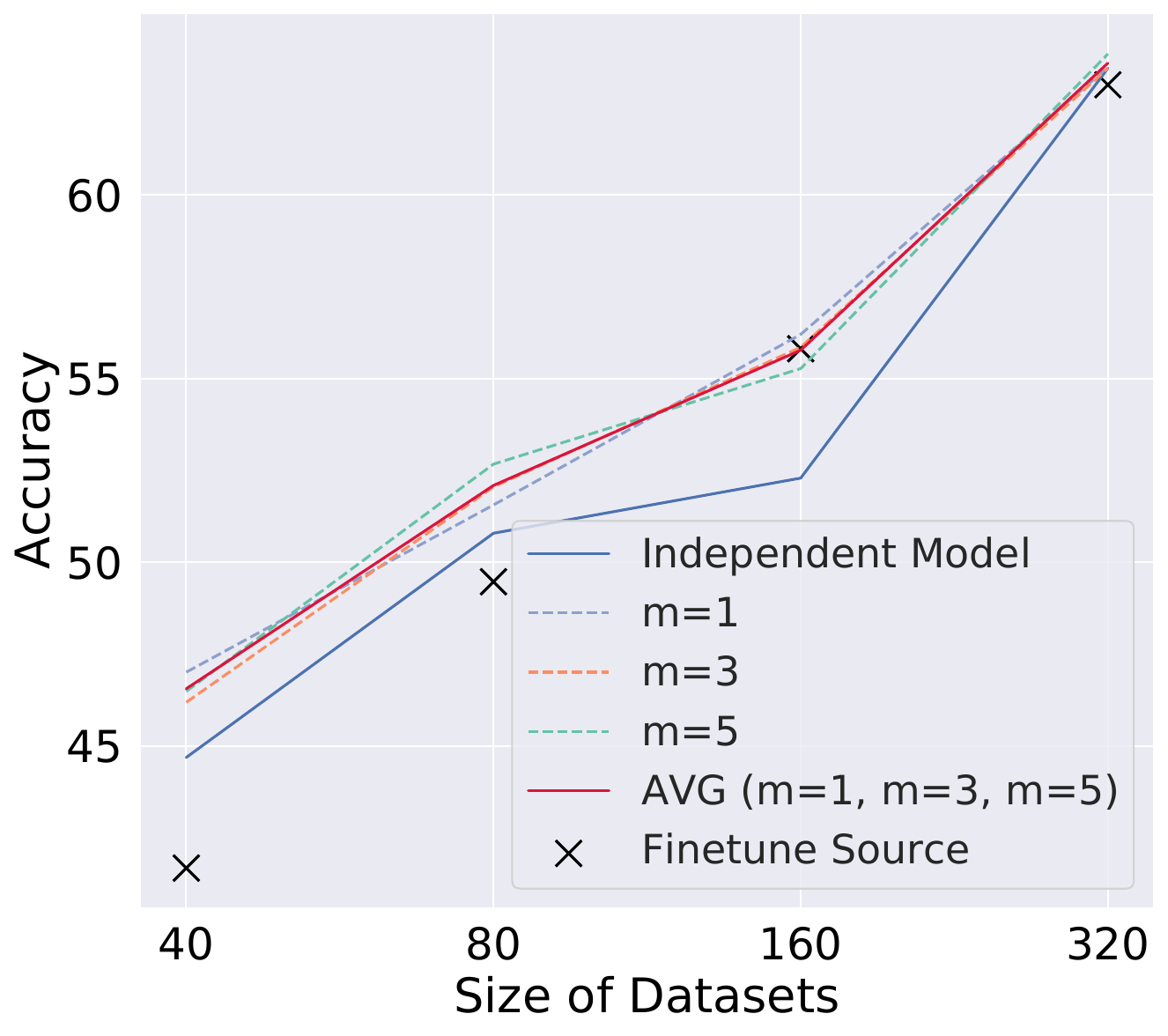}
   \hspace{4mm}
   \includegraphics[width=0.45\textwidth]{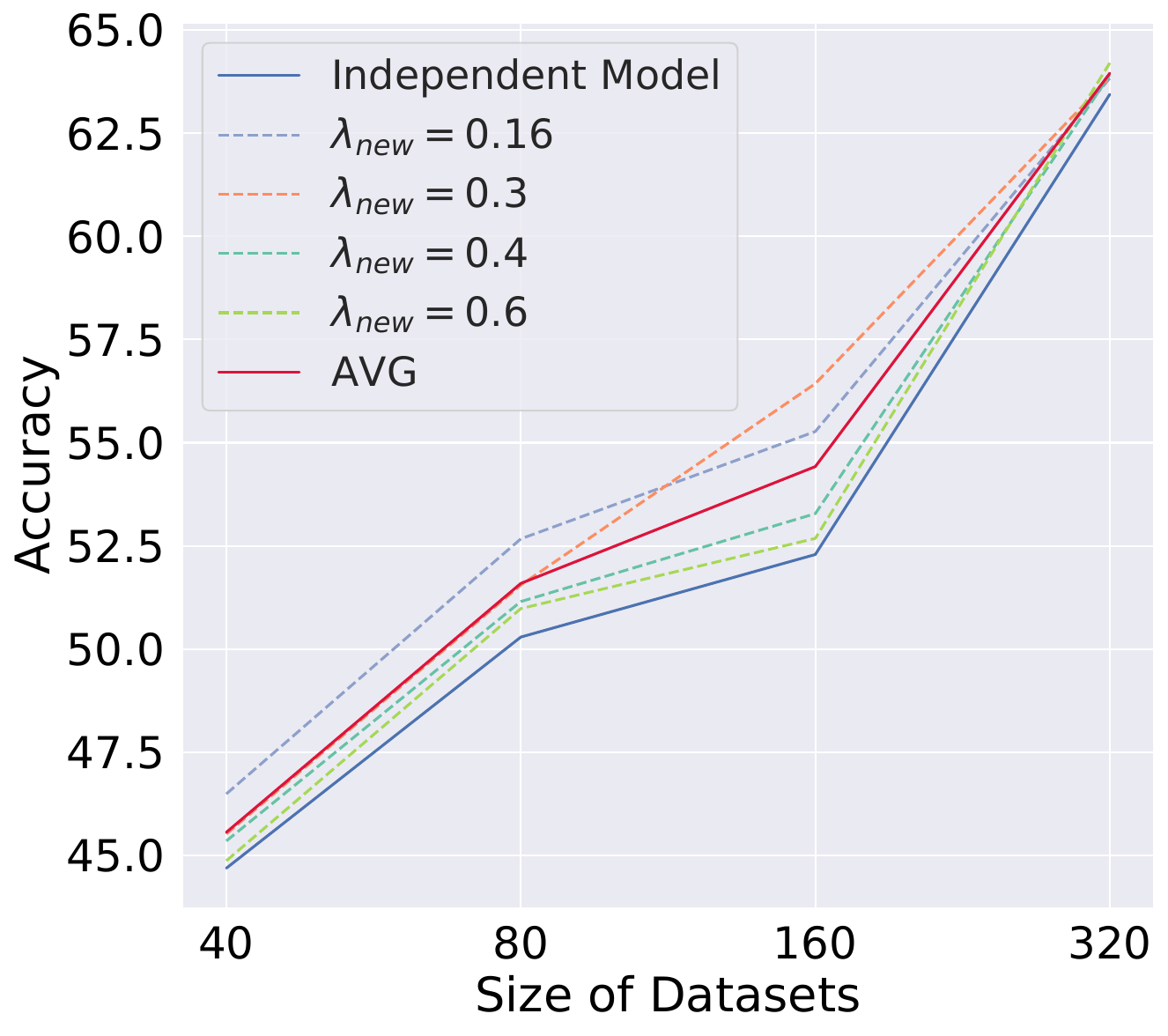}
   \caption{Left: Analysis of the influence of dataset size with respect to the number of source models adapted on \texttt{CIFAR100}. Right: Analysis of the influence of dataset size $e$ with respect to different mixing weight initialization on \texttt{CIFAR100} ($m=5$ for all the experiments). $\lambda_{new}$ on the right panel represents the initial value of the new random initialized modules.}
\label{fig:cifar100}
\end{figure}

\subsection{White-box Scenario}

\paragraph{CIFAR100} \label{sec:CIFAR100} We create a set of 40 tasks with \texttt{CIFAR100}. Different tasks in the collection may share some of the same classes. However, we ensure that the overlapping classes do not have the same data samples by splitting data for each class into two even parts and only allowing each class to exist in tasks twice; this allows us to create a scenario where each task is private and has unique data samples. Details of each task are shown in the Appendix \ref{A.2.1}. We run each experiment 3 times. For each run, we randomize the list of 40 tasks and pick the first ten of the list as the starting source task pool. For the rest of the tasks, $11-40$, we proceed as follows. First, for target task $11$, we randomly select a subset of size $e=\{40, 80, 160, 320\}$. With each subset, we select $m=\{1, 3, 5\}$ related source tasks from the pool to train our new module-mixing model.
Before moving to target task $12$, we train an independent model (via EFT) for task $11$ with all available data. This is then repeated for all the other tasks, $12-40$. Performance is reported for different values of $m$ and $e$ averaged over tasks $11-40$.
This is done to obtain results for target tasks of different sizes while making sure all source models are trained with sufficient data.

\paragraph{Influence of the Number of Source Models}
It is important to note that each task has a different optimal selection of $m$ since the number of related tasks with positive transfer is different for each task. However, for the sake of simplification, we report the results with a fixed number of  source models (with $m=1, 3, 5$) for each task. In Figure \ref{fig:cifar100} (Left), we observe that with all of the chosen $m$, we obtain better results than directly finetuning a source model and training from scratch. Moreover, a wider gap is observed between the independent model and the average performance of our module-mixing model with varying numbers of source models as the size of the datasets decreases, indicating that one potentially gets better performance gains from our method compared to training an independent model. Further, directly finetuning a source model may result in overfitting, especially when the dataset size is small. Alternatively, our model effectively reduces the chances of overfitting.

\paragraph{Influence of Different Mixing Weight Initialization}
We also show the effect of the mixing weights' initialization on overall performance. We consider all the source tasks equally, hence no extra information on which one is more important from the start. Hence, we initialize their mixing weights with equal values and only change the weight values of the new random initialized modules, {\em i.e.}, $\lambda_{new}=\lambda_{m+1,J+1}$. For simplicity, all layers share the same set of weights at initialization. In Figure \ref{fig:cifar100} (Right), for $m=5$, we observe that the average accuracy for different mixing weight initialization is always better than the independent model, albeit with particular initialization values one may get better performance than the average.

\paragraph{Office-31} \texttt{Office-31} consists of 31 categories of images originating from 3 domains: Amazon, DSLR and Webcam, including common objects in everyday office settings. Webcam consists of 795 low-resolution images with noise; the Amazon domain has 2817 images from online merchants; DSLR images are captured with a digital camera, and all 423 images are of high resolution and low noise. With this dataset, we show how our model behaves to covariate shift, when all the tasks have sufficient data and share the same labels. 

In this context, Finetune Source becomes a more competitive baseline now that it does not suffer from serious overfitting issues. From Table~\ref{table:office-31}, our model $m=1$ guarantees that it has better if not equal performance compared to finetuning the same source model. Moreover, we achieve a $7.75\%$ performance increase compared to Finetune Source when adapting to both two source models training with DC loss. We also see that on average, learning from more source models that are trained on a related domain yields better positive transfer, specifically, we observe a $2.78\%$ increase from going from $m=1$ to $m=2$ when training with DC loss. Since MCW is designed as a fast-to-adapt and lightweight model that does not introduce too many trainable parameters, it loses its advantage in this setting when compared to other finetuning models, and even multi-source SVM when the source models are trained on closely related domains. As for cross-dataset soup and ensemble methods like model stacking, they work better when all the models are trained on the same datasets (or different splits of the same dataset). 

\begin{table}[t]
\caption{Accuracy comparison on \texttt{Office-31} in the white-box scenario. The highest accuracy is marked in bold. We also compare the results training with and without distance correlation loss for our model.}
\begin{center}
\begin{tabular}{lcccc}
Method & $A, D \rightarrow W$ & $A, W \rightarrow D$ & $W, D \rightarrow A$ & AVG\\
\midrule
Model stacking & 58.75 & 68.62 & 64.31 & 63.89\\
Cross-dataset soup & 11.88 & 21.78 & 10.84 & 14.83\\
Multi-source SVM & 90.11 & 81.82 & 84.98 & 85.64\\
MCW      & 64.84 & 62.12 & 76.79 & 67.92\\
DATE (SHOT) & 60.74 & 52.74 & 20.43 & 44.63 \\
DECISION & 59.22 & 50.42 & 21.35 & 43.66 \\
Independent  & 89.24 & 75.62 & 71.73 & 72.34\\
Finetune Source & 91.25 & 88.24 & 82.69 & 87.39\\
[5pt]
$m=1$ (w/o $\CL_{DC}$) & \textbf{97.50} & 88.24 & 91.52 & 92.42\\
$m=2$ (w/o $\CL_{DC}$) & 95.00 & \textbf{94.12} & 91.17 & 93.43\\
$m=1$ (w/ $\CL_{DC}$)& 95.00 & 90.20 & 91.87 & 92.36\\
$m=2$ (w/ $\CL_{DC}$)& \textbf{97.50} & \textbf{94.12} & \textbf{95.05} & \textbf{95.14}\\
\end{tabular}
\label{table:office-31}
\end{center}
\end{table}

\begin{table}[t]
\caption{Accuracy comparison on \texttt{CTrL} in the white-box scenario. For our model, besides showing the effect of the distance correlation loss, we also show the importance of a hyperparameter grid search when the model pool is filled with models that are trained on very different source datasets.}
\begin{center}
\begin{tabular}{lcccc}
 Method & $m=1$ & $m=3$ & $m=5$ & AVG\\
\midrule
Model stacking & 42.53 & 34.02 & 34.31 & 36.95\\
Cross-dataset soup & 47.73 & 43.83 & 39.43 & 43.66\\
Multi-source SVM & 44.34 & 28.21 & 26.59 & 33.08\\
MCW  & 42.73 & 45.02 & 54.09 & 47.28\\
Independent & -& - & - & 45.77\\
Finetune Source & - & - & - & 54.26\\
DATE (SHOT) (knn) & 42.45 & 43.72 & 42.84 & 43.00 \\
DECISION (knn) & 42.56 & 45.24 & 45.68 & 44.49 \\
[5pt]
Multi-source SVM (knn) & \textbf{68.48} & 56.96 & 48.33 & 57.92\\
MCW (knn)  & 66.59 & \textbf{72.89} & \textbf{72.91} & \textbf{70.80}\\
Ours (w/o $\CL_{DC}$) & 56.40 & 55.13 & 52.09 & 54.54 \\
Ours (w/$\CL_{DC}$) & 56.48 & 55.35 & 52.79 & 54.87\\ 
Ours (w/$\CL_{DC}$ grid search) & 67.26 & 70.24 & 68.71 & 68.73\\
\end{tabular}
\label{table:ctrl results}
\end{center}
\end{table}

\begin{table}[t]
\caption{Accuracy comparison on \texttt{CIFAR100} in the black-box scenario. For our model, we also compare the results training with and without distance correlation loss.}
\begin{center}
\begin{tabular}{lccccc}
Method & $m=1$ & $m=3$ & $m=5$ & API & AVG\\
\midrule
Model stacking & 71.32 & 70.97 & 70.36 & - & 70.88\\
Cross-dataset soup & 39.71 & 37.17 & 35.93 & - & 37.60\\
Multi-src SVM & 54.76 & 51.48& 47.60 & - & 51.28\\
MCW & 50.46 & 52.75 & 59.34 & - & 54.18\\
DINE & 40.41 & 41.23 & 39.76 & - & 40.47 \\
Independent & - & - & - & - & 70.39\\
[5pt]
Ours (w/o $\CL_{DC}$) & 71.43 & 71.74 & 71.17 & 69.97 & 71.08\\
Ours (w/ $\CL_{DC}$) & \textbf{72.03} & \textbf{71.82} & \textbf{71.46} & \textbf{71.80}& \textbf{71.78}\\
\end{tabular}
\label{table:cifar100 blackbox}
\end{center}
\end{table}

\paragraph{CTrL}
The $S^{long}$ stream from \texttt{CTrL} is a collection of 100 tasks created from five well-known computer vision datasets: \texttt{CIFAR10}, \texttt{CIFAR100}, \texttt{SVHN} \citep{netzer2011reading}, \texttt{MNIST} \citep{lecun2010mnist}, and \texttt{Fashion-MNIST} (short as \texttt{FMNIST}) \citep{xiao2017fashion}. Each task in $S^{long}$ is constructed by first randomly selecting a dataset, then five classes of the chosen dataset, and finally, a large task (containing 5000 training samples) or a small task (containing 25 training samples). During tasks $1-33$, the fraction of small tasks is $50\%$; this increases to $75\%$ for tasks $34-66$, and to $100\%$ for tasks $67-100$. More details of the dataset can be found in the Appendix \ref{A.2.2}. In our experiments, we use the first 60 tasks as our starting pool of source models and gradually add newly trained models into the collection. We report the average test accuracy on the last 40 tasks. This is a challenging problem not only because it simulates a realistic setting that starts with a large collection of pre-trained tasks for repurposing, but also because the last 40 tasks have only 25 training samples each, which could cause serious overfitting problems. 
We perform a simple grid search on hyperparameters (learning rate and weight decay) and compare them to having the same settings for all tasks. Grid search and other experimental details are in the Appendix \ref{A.2}. Results show that grid search is necessary when tasks come from very different datasets. 
In Table~\ref{table:ctrl results}, we also show results of an extension of MCW and multi-source SVM with our $k$-NN source model sampler. The results show the effectiveness of our $k$-NN source model sampler, helping to achieve a $23.52\%$ increase for MCW compared to learning from randomly sampled source models. However, we still suffer from the overfitting problem mainly on tasks created with \texttt{SVHN}, which explains the gap between our method and MCW with $k$-NN sampler. 
\subsection{Black-box Scenario}
Results on \texttt{CIFAR100} (with all data in each task) are shown in Table \ref{table:cifar100 blackbox}. We also used $k$-NN to select source models for baselines. The results show that we can still benefit from such a simple approach. Additional experiment results on other datasets in this setting can be found in the Appendix \ref{A.6.6}.

\begin{figure}[ht]
\centering
\includegraphics[width=0.3\linewidth]{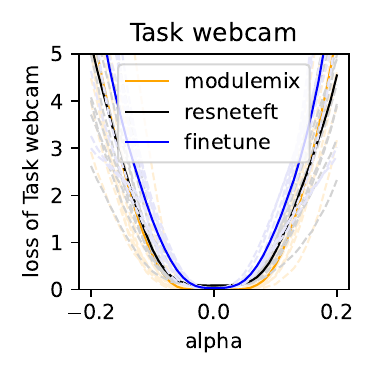}
\includegraphics[width=0.3\linewidth]{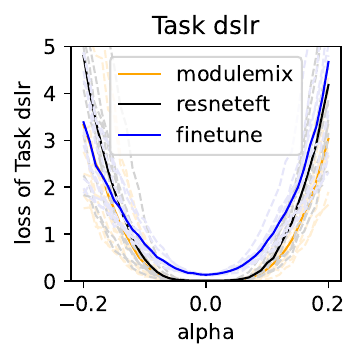}
\includegraphics[width=0.3\linewidth]{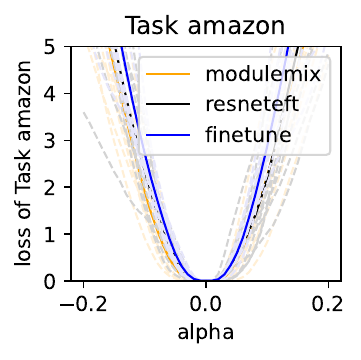}
\caption{Comparison of weight loss landscapes. The dotted lines shows the loss created with the random directions given the trained model weights.}
\label{fig:loss_landscape}
\end{figure}

\subsection{Insights Behind Module-Mixing Framework Design}
\cite{deng2021flattening, Liebenwein2021SparseFP, Cha2021SWADDG, Wortsman2022ModelSA} have shown that a model with a flatter loss landscape contributes to the generalization of models in domain adaptation and continual learning, thus is more robust to overfitting compared to directly finetune the source model. A similar idea as ours proposed in \cite{Wortsman2022ModelSA} showed that by averaging weights of multiple finetuned models with the same initialization, the model’s optimum falls in a flatter loss/error landscape with overall lower loss/error. Based on \cite{Li2017VisualizingTL}, the sharpness of loss landscapes correlates well with generalization error. 

To show that the module-mixing framework has a better generalization ability and more robust to overfitting than directly finetuning source models, in Figure~\ref{fig:loss_landscape} we show the weight loss landscape of the independent model (resneteft), finetune, and our module-mixing model $(m=1)$ trained on all data from three domains in \texttt{Office31}. We follow the procedure from \cite{deng2021flattening} and create ten random directions given the trained model weights. The solid line is the average over the results from the ten random directions. We can see from the plots that module-mixing model has a flatter loss landscape than the other methods, which explains why the model alleviates the effects of potential overfitting more than directly finetuning the model.

We also provide another plot for module-mixing under black-box setting with limited target data in the Appendix \ref{A.6.1}. In the figure, we compare the module-mixing model with the Finetune Source model trained on webcam with only $20\%$ of its data. Same conclusions can be drawn from this setting.

\begin{figure}[t]
\centering
\includegraphics[width=0.6\linewidth]{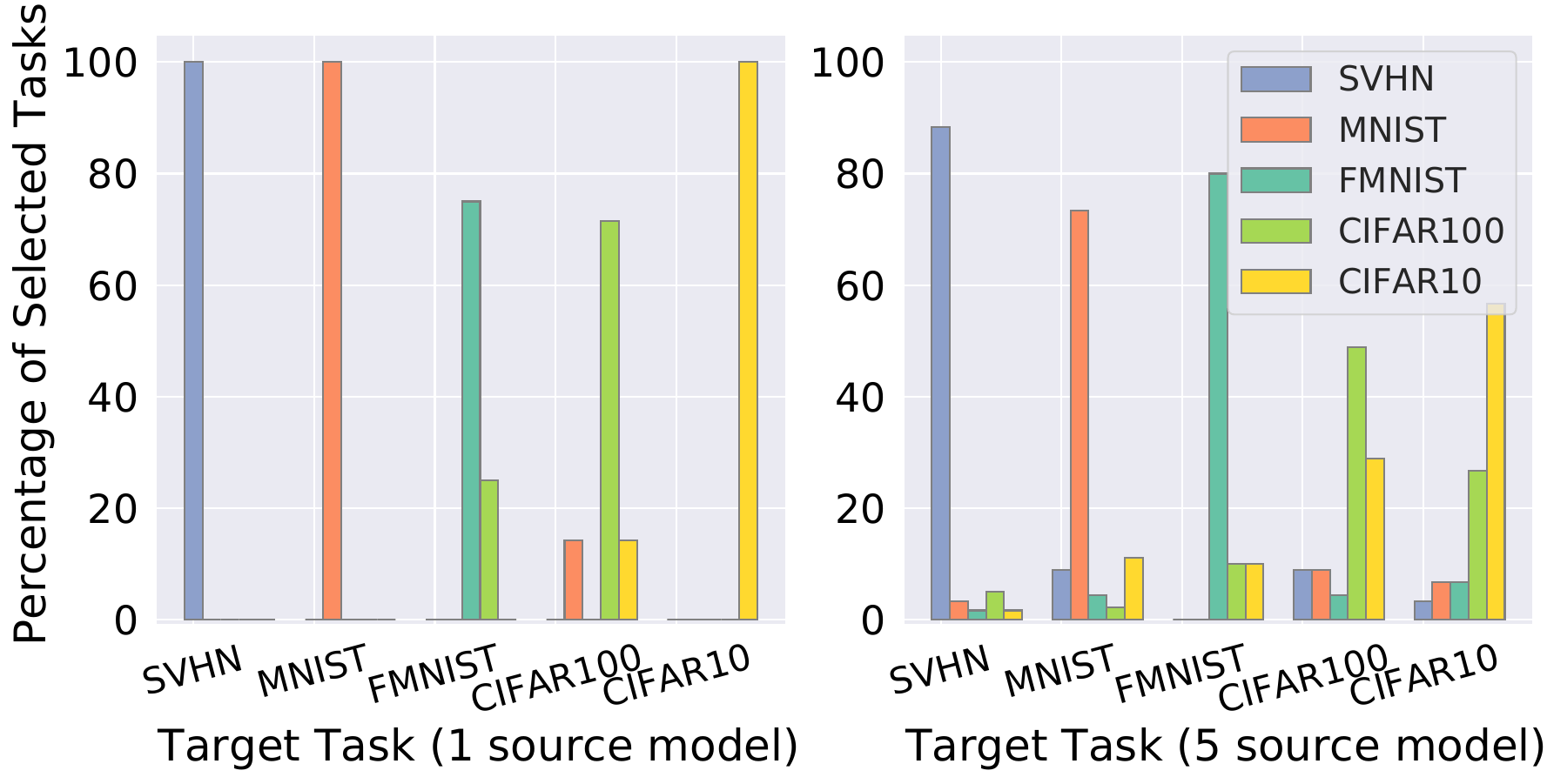}
\includegraphics[width=0.35\linewidth]{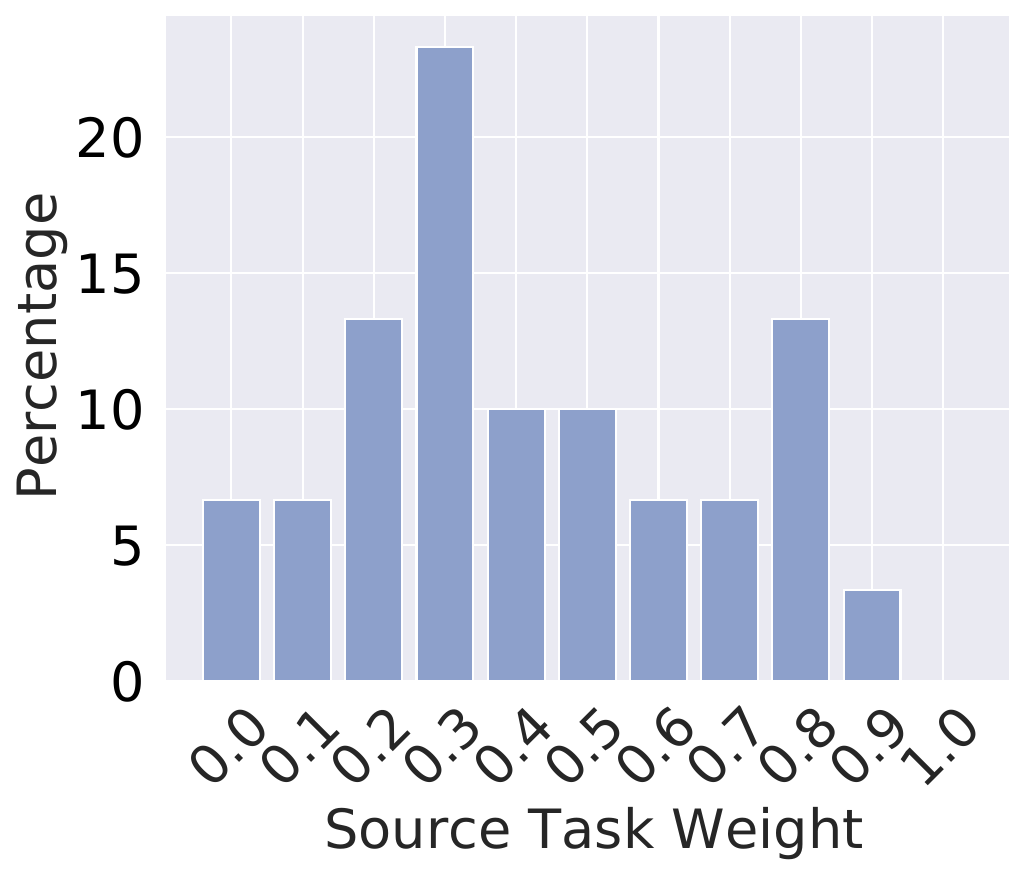}
\caption{Selection of $m=1$ (Left) and $m=5$ (Middle) related tasks via $k$-NN. The $x$-axis shows the membership of the target tasks, while the $y$-axis shows the proportion of selected source tasks for each membership. Right: Module-mixing results for $m=1$. The $x$-axis is the mixing weight value set for the source model and the $y$-axis is the percentage of tasks in \texttt{CIFAR100} for which the highest accuracy is obtained at a given source weight setting.
}
\label{fig:ctrlknnselection}
\end{figure}

\subsection{Ablation Study}\label{ablation}
We construct a module-mixing model with the top-$1$ source model selected via $k$-NN and a target model using only \eqref{eq:feat_comb}. In this setting, the mixing weight $\lambda_1$ for the source model features is set as a value in $\{0, 0.1, \ldots, 0.9, 1\}$, and only the target model's modules and the classification head are trained. Thus, an independent model is trained when $\lambda_1=0$, whereas only the classification head of a pre-trained source model is trained when $\lambda_1=1$. We test on our 30 \texttt{CIFAR100} tasks.
For each task, one setting for $\lambda_1$ will achieve the highest test accuracy.
Each bar in Figure \ref{fig:ctrlknnselection} (Right) shows the percentage of tasks that reached peak accuracy at the given task weight setting. We see that the optimum results are mostly obtained (93\%) when the mixing weight is non-zero for either source and target models, thus justifying our model design. 

\begin{table}[ht]
\small
\caption{Ablation Study on \texttt{CTrL} dataset when $m=1$.}
\begin{center}
\begin{tabular}{lc}
Method & Accuracy\\
\midrule
Independent model & 45.77\\
Random (only \eqref{eq:feat_comb}) &  45.88\\
$k$-nearest neighbors (only \ref{eq:feat_comb}) & 55.68\\
Random (both \eqref{eq:feat_comb} and \eqref{eq:param_comb}) & 46.61\\
$k$-nearest neighbors (both \eqref{eq:feat_comb} and \eqref{eq:param_comb}) & {\bf 56.40} \\
\end{tabular}
\label{table: ablation}
\end{center}
\end{table}

In Table~\ref{table: ablation}, we examine the advantages of our source model selection and layer-wise module-mixing design on \texttt{CTrL} (with $m=1$ for adaptation).
Alternatively, we consider selecting source models at random (instead of $k$-NN) and only using the convex combination of features in \eqref{eq:feat_comb} (instead of both features and module parameters).
We see that with random selection and only feature aggregation, we get an insignificant gain of $0.11\%$ relative to the independent model, while with $k$-NN selection and only feature aggregation, the gain is higher at $1\%$. Importantly, the complete approach using \eqref{eq:feat_comb} and \eqref{eq:param_comb} with k-nearest neighbors model ewselection yields a substantial $9.79\%$ gain relative to that with random source model selection.

\begin{table}[t]
\caption{\texttt{TinyImageNet} in the white-box scenario. Selected task names are colored for clarity.}
\begin{center}
\small
\begin{tabular}{clccccccc}
 & & \color{Thistle}{Task1} & \color{SkyBlue}{Task2} & \color{SeaGreen}{Task3} & \color{YellowOrange}{Task4} & \color{Red}{Task5} & \color{Violet}{Task6} & AVG \\
 \midrule
 & Independent model & 34.55 & 32.64 & 35.07 & 32.47 & 33.16 & 33.25 & 33.52\\
& Finetune source & 34.38 & 31.77 & 33.51 & 32.73 & 33.59 & 32.90 & 33.15 \\
[5pt]
\multirow{6}{*}{$m=2$} & Model stacking & 31.03 & 31.86 & 35.33 & 31.43 & 30.64 & 28.91 & 31.53\\
& Cross-dataset soup & 28.65 & 32.20 & \textbf{37.24} & \textbf{34.20} & 33.51 & 32.73 & 33.09\\
& Multi-source SVM & 28.64 & 25.44 & 25.76 & 24.72 & 27.04 & 23.76 & 25.89 \\
& MCW  & 21.76 & 22.32 & 22.40 & 21.68 & 22.16 & 21.68 & 22.00\\
& DATE (SHOT) & 21.24 & 22.76 & 21.49 & 22.36 & 21.66 & 21.90 & 21.90 \\
& DECISION & 21.34 & 21.93 & 21.36 & 21.58 & 22.46 & 20.96 & 21.60\\
[5pt]
\multirow{6}{*}{$m=1$} & LogME & 35.16 & \textbf{34.80} & 35.36 & 32.90 & 35.07 & 33.28 & 34.43 \\
& Selected task & \color{SkyBlue}{Task2} & \color{Thistle}{Task1} & \color{SkyBlue}{Task2} & \color{Red}{Task5} & \color{YellowOrange}{Task4} & \color{Red}{Task5}\\
& LEEP & 35.16 & 33.59 & 35.36 & 32.90 & 35.07 & 34.08 & 34.36 \\
& Selected task & \color{SkyBlue}{Task2} & \color{SeaGreen}{Task3} & \color{SkyBlue}{Task2} & \color{Red}{Task5} & \color{YellowOrange}{Task4} & \color{SeaGreen}{Task3}\\
& Ours & 35.16 & 33.59 & \textbf{36.26} & 32.90 & \textbf{35.07} & \textbf{35.40} & \textbf{34.73} \\
& Selected task & \color{SkyBlue}{Task2} & \color{SeaGreen}{Task3} & \color{YellowOrange}{Task4} & \color{Red}{Task5} & \color{YellowOrange}{Task4} & \color{YellowOrange}{Task4}\\
[5pt]
\multirow{9}{*}{$m=2$} & LogME & 35.00 & 34.08 & 33.68 & 33.52 & 34.96 & 33.36 & 34.10 \\
& \multirow{2}{*}{Selected tasks} &\multicolumn{1}{l}{\color{SkyBlue}{Task2}} & \multicolumn{1}{l}{\color{Thistle}{Task1}} & \multicolumn{1}{l}{\color{SkyBlue}{Task2}} & \multicolumn{1}{l}{\color{SeaGreen}{Task3}} & \multicolumn{1}{l}{\color{YellowOrange}{Task4}} & \multicolumn{1}{l}{\color{YellowOrange}{Task4}} \\
& & \multicolumn{1}{l}{\color{Violet}{Task6}} & \multicolumn{1}{l}{\color{SeaGreen}{Task3}} & \multicolumn{1}{l}{\color{YellowOrange}{Task4}} & \multicolumn{1}{l}{\color{Red}{Task5}} & \multicolumn{1}{l}{\color{Violet}{Task6}} & \multicolumn{1}{l}{\color{Red}{Task5}} \\
& LEEP & 35.00 & 34.08 & 33.68 & 33.52 & 34.96 & 34.96 & 34.37 \\
& \multirow{2}{*}{Selected tasks} &\multicolumn{1}{l}{\color{SkyBlue}{Task2}} & \multicolumn{1}{l}{\color{Thistle}{Task1}} & \multicolumn{1}{l}{\color{SkyBlue}{Task2}} & \multicolumn{1}{l}{\color{SeaGreen}{Task3}} & \multicolumn{1}{l}{\color{YellowOrange}{Task4}} & \multicolumn{1}{l}{\color{SeaGreen}{Task3}} \\
& & \multicolumn{1}{l}{\color{Violet}{Task6}} & \multicolumn{1}{l}{\color{SeaGreen}{Task3}} & \multicolumn{1}{l}{\color{YellowOrange}{Task4}} & \multicolumn{1}{l}{\color{Red}{Task5}} & \multicolumn{1}{l}{\color{Violet}{Task6}} & \multicolumn{1}{l}{\color{Red}{Task5}} \\
& Ours & \textbf{36.40} & 34.08 & 33.68 & 33.52 & 34.96 & 33.84 & 34.41 \\
& \multirow{2}{*}{Selected tasks} &\multicolumn{1}{l}{\color{SkyBlue}{Task2}} & \multicolumn{1}{l}{\color{Thistle}{Task1}} & \multicolumn{1}{l}{\color{SkyBlue}{Task2}} & \multicolumn{1}{l}{\color{SeaGreen}{Task3}} & \multicolumn{1}{l}{\color{YellowOrange}{Task4}} & \multicolumn{1}{l}{\color{Thistle}{Task1}} \\
& & \multicolumn{1}{l}{\color{SeaGreen}{Task3}} & \multicolumn{1}{l}{\color{SeaGreen}{Task3}} & \multicolumn{1}{l}{\color{YellowOrange}{Task4}} & \multicolumn{1}{l}{\color{Red}{Task5}} & \multicolumn{1}{l}{\color{Violet}{Task6}} & \multicolumn{1}{l}{\color{YellowOrange}{Task4}} \\
\end{tabular}
\end{center}
\label{table:tinyImagenet}
\end{table}

To show that our method also applies to more complex tasks, we provide an experiment on \texttt{Tiny-ImageNet}. We use a randomly initialized ResNet-18 as the backbone. The purpose of this setting is to show that the improvement in performance of our model does not rely on a powerful and closely related backbone model since \texttt{Tiny-ImageNet} is a subset of \texttt{ImageNet}. We create 6 tasks with 200 classes in the dataset, with 50 classes for each task. Task 1 through 5 have overlapping classes, while Task 6 has no overlapping with other tasks. Task details are presented in Table \ref{table:tinyImageNet Details} and \ref{table:tinyImageNet Details Continue}, where tasks with overlapping classes have the same color-coding. Since the tasks in this setting all have sufficient training data, the newly initialized task-specific modules for the target task are given higher initial module-mixing weights (0.9) to encourage the learning of target task-specific knowledge.
For each task, we search for top-$m$ most similar tasks in the other 5 tasks. In Table \ref{table:tinyImagenet}, we also provide the source model selection results, shown in ``selected task''. The task names have the same colors as in the header of the table to provide more clarity.
We observe that tasks with overlapping classes are always picked for each task. 
For instance, Task 2 overlaps with Task 1 and 3, so when we build our module-mixing model with the top-$2$ source models to reuse, these tasks are selected.

Furthermore, we compare with LEEP \citep{nguyen2020leep} and LogME \citep{you2021logme} in identifying the most transferable source models. Different model transferability evaluation methods are first used to select the top-$m$ transferable models, and then the selected source models are used to build our module-mixing models to transfer to each target task. As is shown in Table \ref{table:tinyImagenet}, our selection method's performance is on par with the baselines, while having a slight edge over them. We also provided the average results and standard deviation over three runs comparing our method to LogME and LEEP in the Appendix \ref{A.6.2}. 

Moreover, we show how well the $k$-NN algorithm selects source models with \texttt{CTrL}. 
In Figure~\ref{fig:ctrlknnselection}, we show selection results with $m=1$ (Left) and $m=5$ (Middle). Each bar represents the proportion of selected tasks from each dataset for the 40 tasks. When $m=1$, target tasks created with \texttt{SVHN}, \texttt{MNIST}, and \texttt{CIFAR10} always pick source models trained from the same datasets. Also, we only have 4 \texttt{FMNIST} tasks in the experiments, which explains why the selector picks a high $25\%$ of \texttt{CIFAR10} source models. As $m$ increases to $5$, tasks with classes from \texttt{SVHN}, \texttt{MNIST}, \texttt{FMNIST} still almost always pick models created from the same dataset. For \texttt{CIFAR10} and \texttt{CIFAR100}, since they both contain images of natural objects, they are more similar.
\begin{figure}[ht]
\centering
\includegraphics[width=\linewidth]{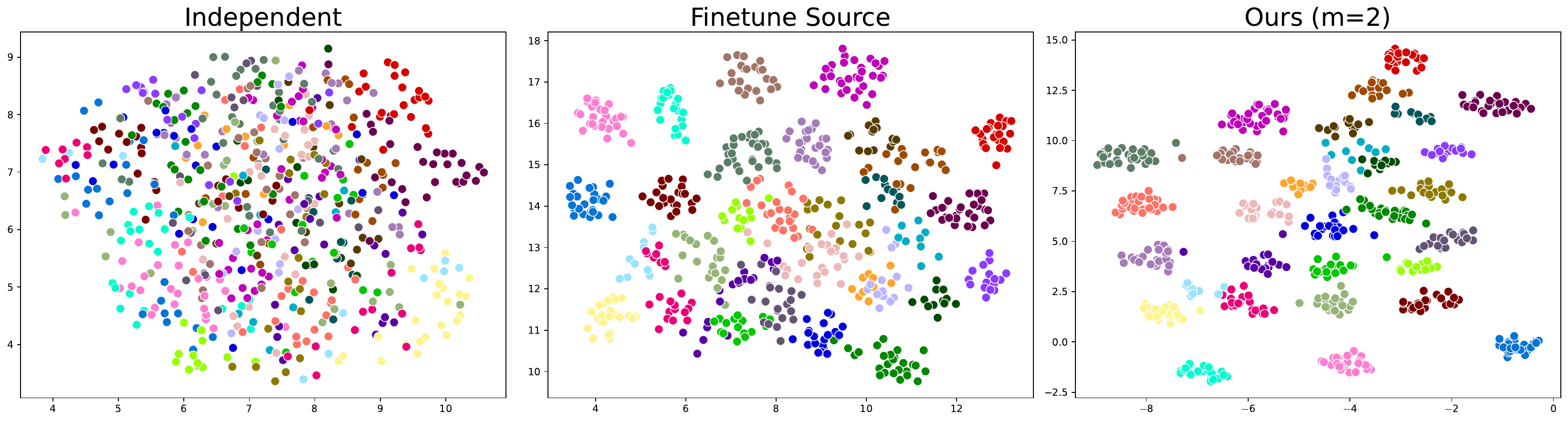}
\caption{Umap plots on the Webcam domain: each color represents the same label across all three panels.}
\label{fig:umap_webcam}
\end{figure}
\subsubsection{UMAP Plots of Office-31 Domains}
Finally, we validate if our model learned meaningful data representations by visualizing the features of training data extracted from the feature extractor of models Independent, Finetune Source and our Module-mixing model, respectively, using UMAP \citep{mcinnes2018umap} as shown in Figure \ref{fig:umap_webcam}. Learned features with our model are relatively more clustered and separated compared to that for the other two methods. We also provide quantitative results examining the separation of the features in Appendix \ref{A.6.4}.
For the DSLR domain, the same conclusions can be drawn, as shown in Figure \ref{fig:umap_dslr} in the Appendix \ref{A.6.3}.
\begin{figure}[ht]
\centering
\includegraphics[width=\textwidth]{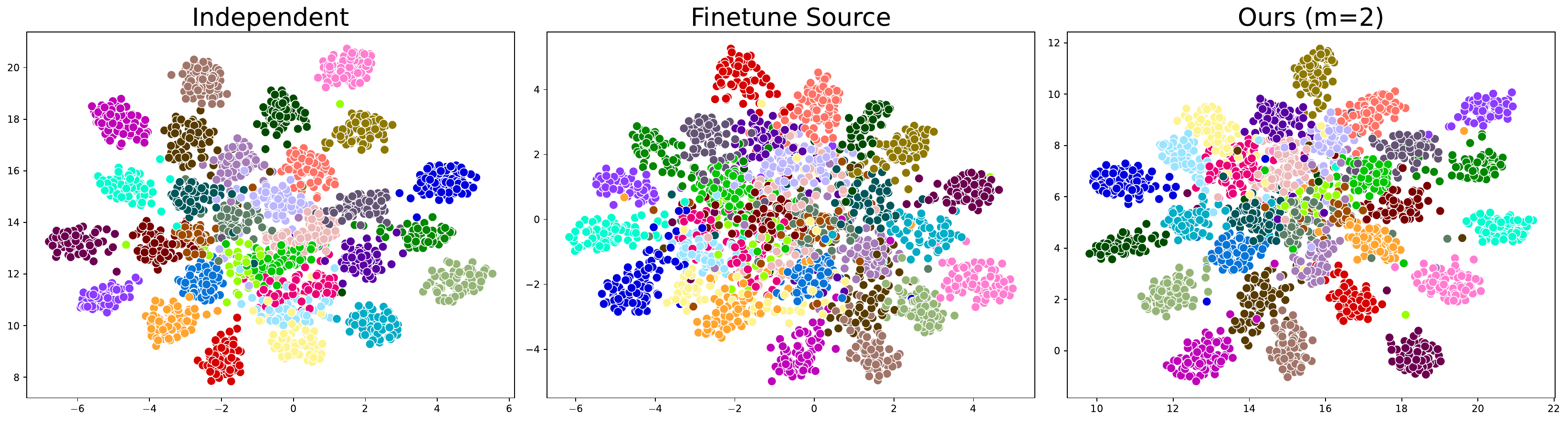}
\caption{Umap plots on the Amazon Domain with training data. Each unique color represents the same label across all three panels.}
\label{fig:umap_amazon}
\end{figure}

\begin{figure}[ht]
\centering
\includegraphics[width=\textwidth]{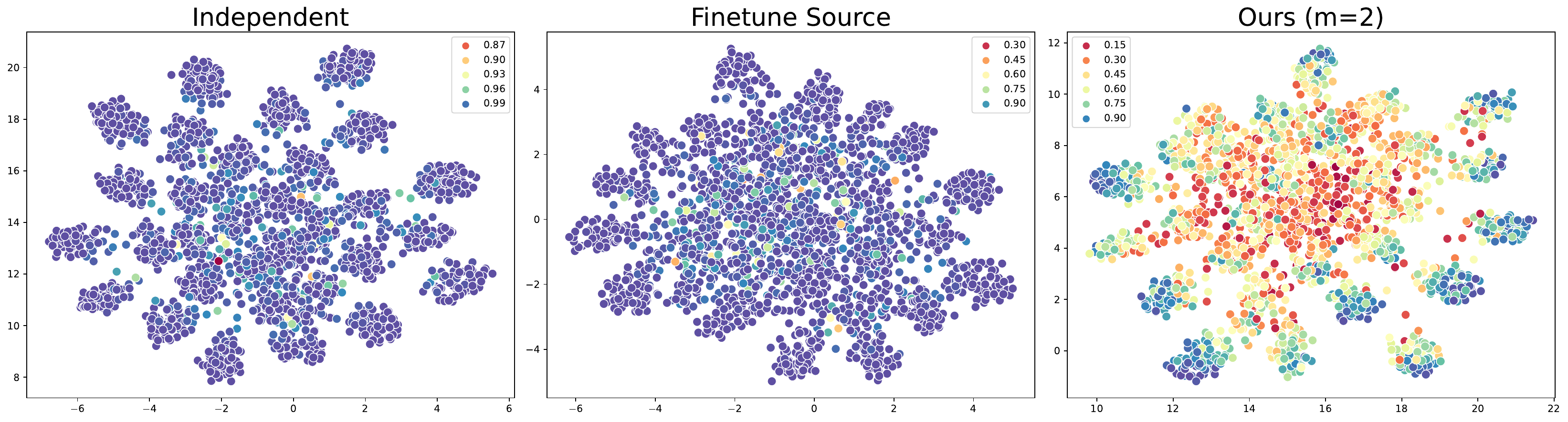}
\caption{Umap plots on the Amazon domain with training data. The colors represent the predicted probability. Independent and Finetune Source have high overall probability for their predictions, while our method does not. This indicates that our method has better generalization.}
\label{fig:umap_amazon_probas}
\end{figure}

\begin{figure}[ht]
\centering
\includegraphics[width=\textwidth]{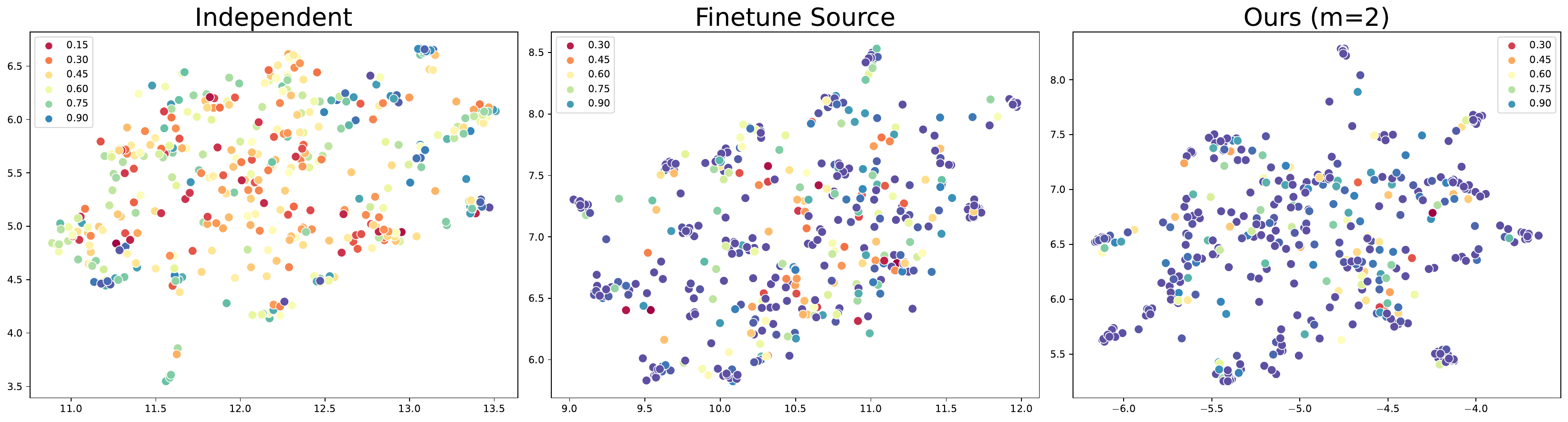}
\caption{Umap plots on the Amazon domain with testing data. The colors represent the predicted probability. Independent and Finetune Source have lower overall probability for their predictions compared to our method's predictions.}
\label{fig:umap_amazon_test}
\end{figure}

For the Amazon domain, in Figure \ref{fig:umap_amazon}, we create scatter plots of the learned UMAP embeddings of the training data, and color samples (points) according to their ground truth labels. All three plots in Figure \ref{fig:umap_amazon} have good clustering patterns. However, if we color-code the training data points by their respective predicted probabilities of outcomes as in Figure \ref{fig:umap_amazon_probas}, we notice that models Independent and Finetune Source are overconfident about their predictions, while our method also has the effect of preventing the model from predicting the labels too confidently during training.
In Figure \ref{fig:umap_amazon_test}, we visualize the extracted features of the test data after UMAP embedding. We see that the Independent model is less confident about the predictions.
While Finetune Source has higher predicted probabilities on the test set, our model has relatively better clustering and is more certain about the predictions compared to Finetune Source. Besides the loss weight landscape analysis, this observation can be seen as another indication that our model has better generalization.

\section{Conclusion}
In this work, we studied an under-explored source-free supervised transfer learning scenario. The proposed framework can efficiently search for related models to adapt to and also can be applied to white-box and black-box source models. To test our module-mixing model and encourage more work in this field, we not only experimented on a classic domain adaptation dataset \texttt{Office-31}, but also crafted new datasets based on existing benchmarks, including \texttt{CIFAR100}, \texttt{CTrL} and \texttt{Tiny-ImageNet}. We showed promising improvements compared to the baselines and examined the properties of each part of our framework. 

Nevertheless, we acknowledge some limitations of the proposed method. So far, we initialize the model's mixing weights with equal values or do a grid search for the best mixing weights. However, it will be interesting to study how to initialize the mixing weights according to different source models' transferability to the target task.
Moreover, we require all models to have task-specific modules of same sizes for the white-box scenario. Therefore, another interesting future work is to see how to build layer-wise module-mixing modules when the task-specific modules are not of the same sizes.

\bibliography{cites}
\bibliographystyle{tmlr}
\clearpage
\appendix
\section{Appendix}

\subsection{Efficient Feature Transformation}\label{A.1}
\cite{Verma2021EfficientFT} proposed efficient feature transformation (EFT) for both 2D convolutional layers and fully connected layers. In the experiments, we apply EFT to convolutional layers in a ResNet18 model. As mentioned in the Background section, given the groupwise convolutional kernels $\omega^s \in \mathbb{R}^{M \times Q \times a}$ and the corresponding $\frac{K}{a}$ groups of feature maps $F$, each group of feature maps $F^s_{ai:(ai+a-1)}$ is convolved with kernels $[\omega^s_{i, 1}, \cdots, \omega^s_{i, a}]$ to form a new group of feature maps $H^s_{ai:(ai+a-1)} \in \mathbb{R}^{M \times Q \times a}$. The formulation of which is shown below:
\begin{align}
    H^{s}_{ai:(ai+a-1)} = [&\omega^s_{i,1} * F_{ai:(ai+a-1)} | \cdots | \\
    \omega^{s}_{i, a}& * F_{ai: (ai+a-1)}], i \in \{0, \cdots, \frac{K}{a}-1\} , \notag
\end{align}
where $|$ is the concatenation operation and $i$ indicates the group of feature maps. Then, all the $H^s_{ai:(ai+a-1)}$ are concatenated to form $H^s$. Same procedures can be applied with the pointwise convolutional kernels $\omega^d \in \mathbb{R}^{M \times Q \times b}$ and the corresponding $\frac{K}{b}$ groups of feature maps $F$:
\begin{align}
    H^{s}_{bi:(bi+b-1)} = [&\omega^s_{i,1} * F_{bi:(bi+b-1)} | \cdots | \\
    \omega^{s}_{i, b}&ss * F_{bi: (bi+b-1)}], i \in \{0, \cdots, \frac{K}{b}-1\} . \notag
\end{align}
Then concatenate $H^d_{ai:(ai+a-1)}$ to form $H^d$. 
Combining $H^d$ and $H^s$ produces the final feature maps $H = H^s + \gamma H^d$, where $\gamma \in \{0, 1\}$, and in this paper we set $\gamma=1$.

By setting $a\ll K$ and $b\ll K$, the amount of trainable parameters per task is substantially reduced. For instance, using a ResNet18 backbone, $a=8$, and $b=16$ results in 449k parameters per new task, which is $3.9\%$ the size of the backbone.
As empirically demonstrated in \cite{Verma2021EfficientFT}, EFT makes model growth efficient in continual learning settings while preserving the remarkable representation power of ResNet.

\subsection{Dataset and Implementation Details} \label{A.2}
\paragraph{Hyperparamter settings} For experiments in the {\em white-box} setting, \emph{i)} with \texttt{Office-31}, all the models are trained for $100$ epochs with batch size $128$; the training starts with a learning rate of $0.001$ and weight decay $1e^{-5}$, and the learning rate decays by $0.1$ after the first 50 epochs. \emph{ii)} For \texttt{CIFAR100}, all the models are trained for $150$ epochs with batch size 128, weight decay of $1e^{-5}$, and a learning rate starts at $0.01$ and decays by $0.1$ after $100$ epochs, except for the tasks with dataset size $40$, which is trained with weight decay $1e^{-4}$. For tasks with dataset size $40$, we also add random horizontal flip data augmentation to all the experiments to further reduce overfitting. \emph{iii)} For \texttt{CtrL}, all the experiments are trained for $100$ epochs with batch size 16. We perform a grid search on learning rate in $\{0, 0.01, 0.003\}$ and weight decay in $\{0, 1e^{-4}, 1e^{-5}\}$, and compare the results to having the same learning rate of $0.01$ and reduced by $0.1$ after 50 epochs, with weight decay $1e^{-4}$. We also add random horizontal flip augmentation, as in the original paper \cite{veniat2020efficient}, to alleviate overfitting.

For experiments in the {\em black-box} setting, the learning rate is set to $0.001$ and reduced by $0.1$ after $50$ epochs, batch size is $128$, and weight decay is $1e^{-5}$ for all datasets. We adopt the scikit-learn package implementation of FastICA to reduce the dimensionality of the feature outputs obtained from the source models. 

\paragraph{Mixing weights settings} In Tables \ref{table:office-31}, \ref{table:ctrl results}, \ref{table:cifar100 blackbox}, \ref{table: ablation}, and \ref{table:tinyImagenet}, uniform initialization is set for the mixing weights in each layer. In the Additional Experimental Results section, results with different mixing weight initialization on \texttt{Office-31} data can be found.

\subsubsection{Details on CIFAR100} \label{A.2.1}
\texttt{CIFAR100} has 100 classes containing 600 images per class, which is further split into 500 training images and 100 test images for each class. We construct a list of 40 tasks $[1, 2, \cdots, 40]$ as in Table \ref{table: cifar100 details} by evenly splitting each class in \texttt{CIFAR100} into 300 images, which means each class in the 40 tasks has 250 training images and 50 test images. For each task, we randomly select a subset of size $e=\{40, 80, 160, 320\}$ out of the training images to train the module-mixing model, with $10\%$ of the selected training images observed as a validation set. However, for training with different subsets, the test set remains the same, which contains 250 images.

\subsubsection{Details on CTrL} \label{A.2.2}
Each task in the $S_{long}$ dataset consists of training, validation, and test datasets. All images have been rescaled to 32x32 pixels in RGB color format, and normalized per-channel using statistics computed on the training set of each task. Data augmentation is performed by using random crops (4 pixels padding and 32x32 crops) and random horizontal flips.

The details of the tasks used in this paper can be found in Tables \ref{table: ctrl slong 1} and \ref{table: ctrl slong 2}. Note that though we used the same random seed (343008097) to generate the dataset as in the original paper, the generated tasks with our hardware have different classes and dataset sizes from the original paper \cite{veniat2020efficient}.

\paragraph{Hyperparameter grid search} In the experiment shown in Table~\ref{table:ctrl results}, learning rates $\{10^{−2}, 10^{−3}\}$, weight decay strengths $\{0, 10^{−5} , 10^{−4}\}$ and weight initializations $\lambda_{new} \in \{0.001, 0.1, 0.2, 0.3, 0.4, 0.5, 0.6, 0.7, 0.8, 0.9, 0.997\}$ are considered (0.001 and 0.997 are used to avoid setting weights to zero). For each $\lambda_{new}$, other mixing weights in the same layer are set to a same value $(1-\lambda_{new})/m$, with $m$ being the number of selected source models. We select the hyperparameter combination that produces the best validation accuracy.

\subsubsection{Details on Office31}
For each domain in Office31, we randomly select $80\%$ of its data as training set, $10\%$ as validation set, and $10\%$ as test set.  Data augmentation is performed by using random crops and random horizontal flips. All images have been rescaled to $256 \times 256$ and then cropped to $224 \times 224$.

\subsection{Ablation Study on the \texorpdfstring{$a/b$}{} Parameter Settings for EFT} \label{A.3}
a=8 and b=16 are the optimal parameter choices according to the original EFT paper \cite{Verma2021EfficientFT} for \texttt{CIFAR100}. We adopted them for our experiments on \texttt{CIFAR100} and \texttt{Office-31}, where the number of samples for each task is relatively sufficient.

We set $a=2$ and $b=1$ for \texttt{CtrL} to reduce the overfitting problem on training source models with limited data since most of the source tasks have very limited data, we reduce the number of training parameters accordingly.

We also tried different $a / b$ settings for \texttt{CtrL}, and $a=2 / b=1$ gives the best performance for independent models. Below are some examples for training independent models with different $a / b$ settings on tasks $60$ to $100$, which all have a limited data size. Weight decay is set to $1e-4$ and learning rate $0.01$ for each setting, and all the experiments are run with the same three random seeds. 

\begin{table}[ht]
\caption{Influence of the number of source models on performance for \texttt{CIFAR100}.}
\begin{center}
\small
\begin{tabular}{ccccc}
 & a=1/b=1 & a=2/b=1 & a=1/b=2 & a=2/b=2\\
\midrule
\makecell{Average test accuracy with \\ three random seeds} & 43.2/44.8/43.9 & 47.2/46.1/45.9 & 44.4/45.3/44.9 & 46.1/45.5/47.2 \\
Avg of three runs & 44.0 & \textbf{46.4} & 44.9 & 46.3\\
\end{tabular}
\end{center}
\label{table:a/b settings}
\end{table}

\subsection{Computational Complexity Analysis for Source Model Selection} \label{A.4}
The computational cost of model selection lies in two parts, one comes from extracting features using all the source models, the other comes from k-NN selection.

Time comsumption for feature extraction depends on the complexity of the source models and the number of points used for model selection. Since the source models’ complexity is predetermined, we can reduce the number of points used for model selection when the number of candidate models is very large.

Given its non-parametric property, k-NN selection is relatively computational efficient. However, here we provide its’ time/space complexity. Given
\begin{itemize}
    \item n: number of points used for model selection
    \item d: feature dimensionality
    \item k: number of neighbors that we consider for voting
\end{itemize}

The time and space complexity for k-NN are
\begin{itemize}
    \item Training time complexity: O(1)
    \item Training space complexity: O(1)
    \item Prediction time complexity: O(k * n * d)
    \item Prediction space complexity: O(1)
\end{itemize}

There is no need for training since all computation is done during prediction. To shorten searching time, two main things can be done. (since feature dimensionality $d$ is predetermined by the source models)

\paragraph{Reduce the value of $n$} We can use a small validation set instead of all the points in the task for model selection. This is also what we did with all the experiments in the paper.

\paragraph{Reduce the value of $k$} Although using a larger $k$ improves the accuracy of model selection, it can increase the time for searching through the model pool. We generally prefer a smaller $k$ value. However, we need to keep in mind the accuracy / efficiency trade-off. Luckily, we found that $k=5$ (or other relatively smaller k value) can already provide a good performance without sacrificing computational cost.

\subsection{Ablation Study on the Setting of \texorpdfstring{$k$}{} for \texorpdfstring{$k$}{}-NN} \label{A.5}
We study how different choices of $k$ for $k$-NN algorithm influence the selection results for source models using the \texttt{CTrL} dataset. We plot the membership selection results in Figure~\ref{fig:k setting for knn}. We can see that the selected source models are more related to the target tasks with a larger value for $k$.
\begin{figure}[ht]
\centering
\includegraphics[width=0.55\textwidth]{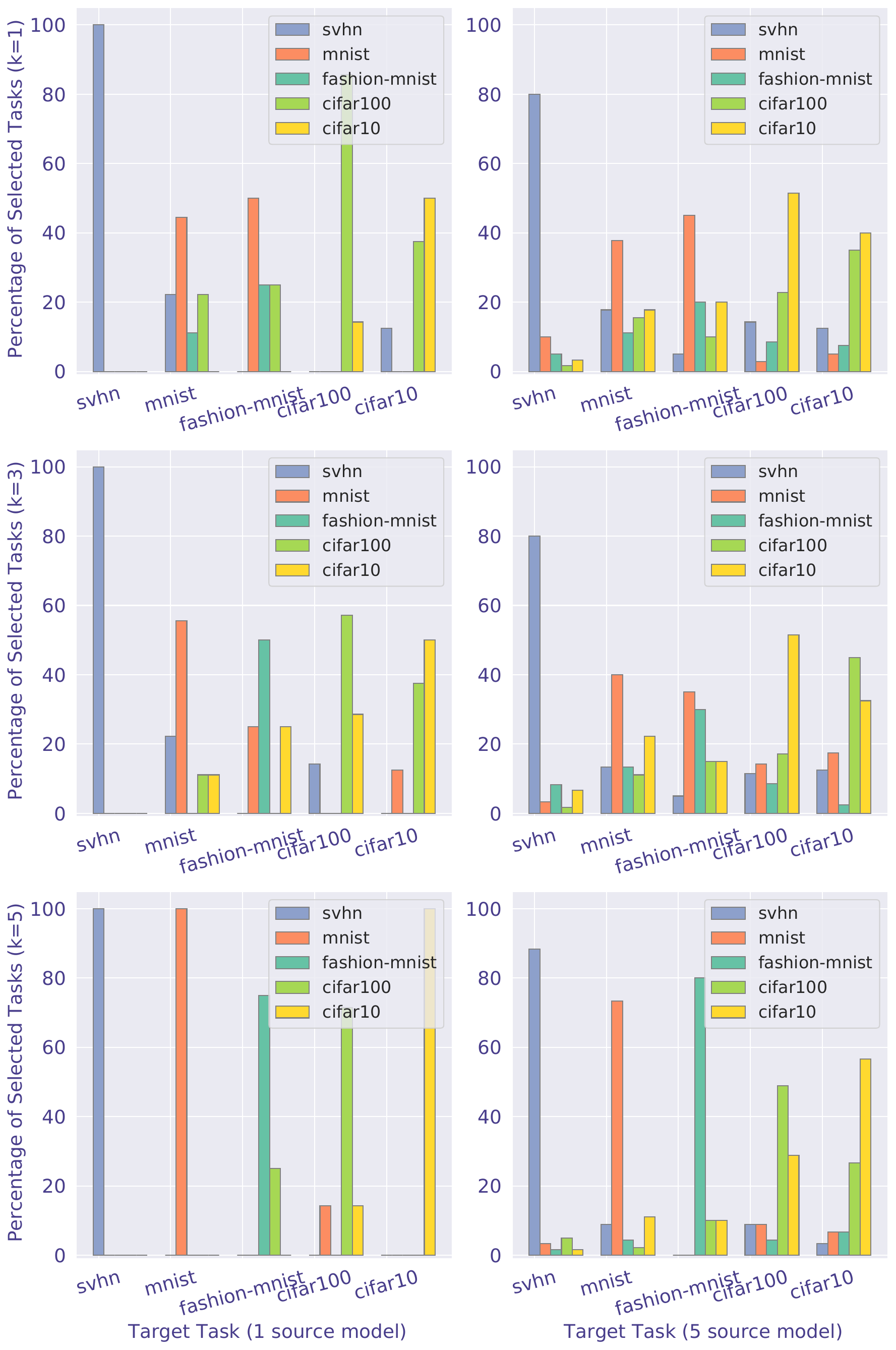}
\caption{Ablation Study on the different setting of $k$ for $k$-NN source model selection. From top to bottom, the results are for when $k=1$, $k=3$ and $k=5$.}
\label{fig:k setting for knn}
\end{figure}

\subsection{Additional Experimental Results} \label{A.6}
\subsubsection{Weight Loss Landscape Plot under Limited Data Setting} \label{A.6.1}
In Figure \ref{fig:loss_limited_data}, we show the weight loss landscape of module-mixing
under black-box setting with limited target data. In the figure, we compare the
module-mixing model with the Finetune Source model trained on webcam with only $20\%$ of its data.

\begin{figure}[H]
\centering
\includegraphics[width=0.4\linewidth]{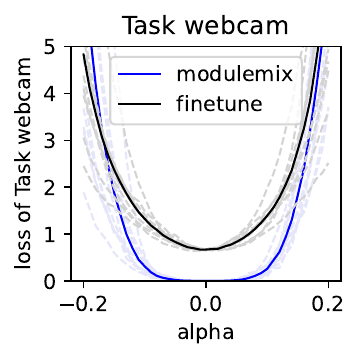}
\caption{Weight Loss Landscape Plot under Limited Data Setting}
\label{fig:loss_limited_data}
\end{figure}
\subsubsection{Comparison to Other Model Transferability Methods} \label{A.6.2}
In Table \ref{table:tinyImagenet three runs}, we show the average and standard deviation over three runs for all three methods (with the same random seeds) to select the most transferable models. 
\begin{table}[ht]
\caption{\texttt{TinyImageNet} in the white-box scenario.}
\begin{center}
\small
\begin{tabular}{clccccccc}
 & & \color{Thistle}{Task1} & \color{SkyBlue}{Task2} & \color{SeaGreen}{Task3} & \color{YellowOrange}{Task4} & \color{Red}{Task5} & \color{Violet}{Task6} & AVG \\
\midrule
\multirow{3}{*}{$m=1$} & LogME & $34.96 \pm 0.52$ & $\textbf{34.47} \pm 0.50$ & $35.42 \pm 0.31$ & $33.04 \pm 0.15$ & $34.79 \pm 0.15$ & $33.79 \pm 0.54$ & 34.41 \\
& LEEP & $35.20 \pm 0.33$ & $33.52 \pm 0.23$ & $35.42 \pm 0.31$ & $33.04 \pm 0.15$ & $34.79 \pm 0.15$ & $34.12 \pm 0.34$ & 34.35 \\
& Ours & $35.20 \pm 0.33$ & $34.06 \pm 0.55$ & $\textbf{36.33} \pm 0.27$ & $33.04 \pm 0.15$ & $\textbf{35.06} \pm 0.22$ & $\textbf{35.56} \pm 0.25$ & \textbf{34.88} \\
[5pt]
\multirow{3}{*}{$m=2$} & LogME & $35.21 \pm 0.20$ & $34.28 \pm 0.27$ & $33.49 \pm 0.24$ & $33.61 \pm 0.09$ & $34.94 \pm 0.25$ & $33.83 \pm 0.43$ & 34.22 \\
& LEEP & $35.21 \pm 0.20$ & $34.28 \pm 0.27$ & $33.49 \pm 0.24$ & $33.61 \pm 0.09$ &  $34.94 \pm 0.25$ & $35.02 \pm 0.32$ & 34.43 \\
& Ours & $\textbf{36.22} \pm 0.34$ & $34.28 \pm 0.27$ & $33.49 \pm 0.24$ & $33.61 \pm 0.09$ & $34.94 \pm 0.25$ & $34.06 \pm 0.20$ & 34.43 \\
\\
\end{tabular}
\end{center}
\label{table:tinyImagenet three runs}
\end{table}
\subsubsection{Additional Umap Plots for Office DSLR Domain} \label{A.6.3}
In Figure \ref{fig:umap_dslr}, learned features with
our model are relatively more clustered and separated compared to that for the other two methods.
\begin{figure}[ht]
\centering
\includegraphics[width=\textwidth]{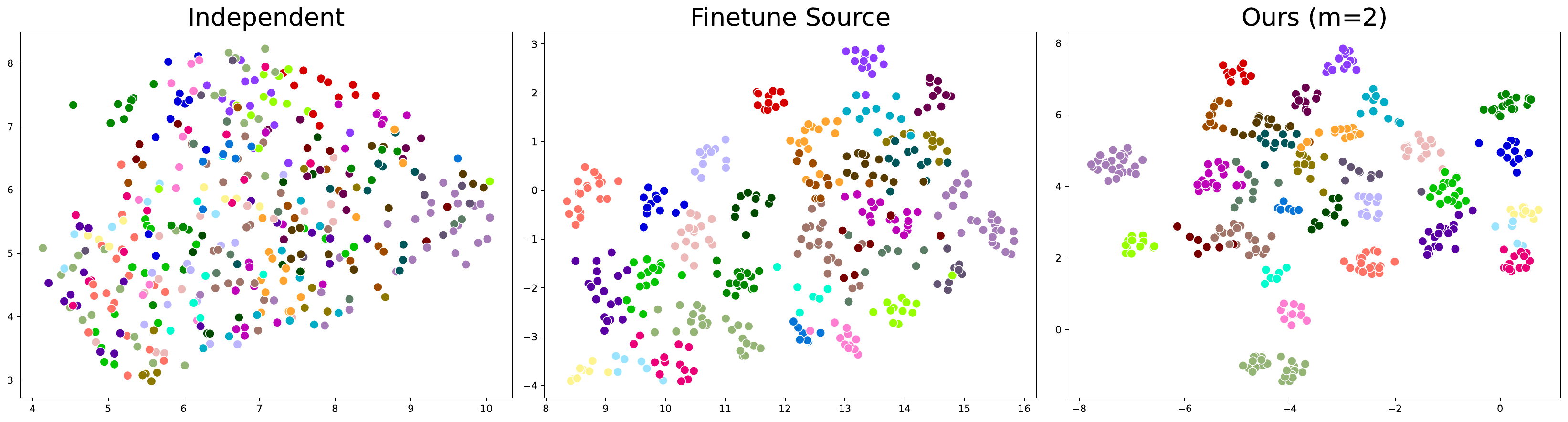}
\caption{Umap plots on the DSLR domain with training data. Each unique color represents the same label across all three plots.}
\label{fig:umap_dslr}
\end{figure}
\subsubsection{Quantitative Analysis on Feature Space} \label{A.6.4}
With the features of training data of each domain extracted from the feature extractor of model Independent, Finetune Source, and our module-mixing module (when $m=2$), we fit KNN models respectively for each feature set. Then we acquire the predicting score with the features of testing set with the trained KNN models respectively. Since the KNN scores are all relatively high and close for the three different methods, we run the experiment for three times and report in Table \ref{table:knn three runs} the average and standard deviation. Our model achieves the highest average KNN score on testing feature set, thus showed the quality of learned features from the quantitative perspective.

\begin{table}[ht]
\caption{KNN scores for Office domains.}
\begin{center}
\small
\begin{tabular}{lccc}
 & Independent & Finetune Source & Ours (m=2) \\
\midrule
Webcam & $0.975 \pm 0.010$ & $0.971 \pm 0.012$ & $\textbf{0.983} \pm 0.012$ \\
Amazon & $0.920 \pm 0.006$ & $0.909 \pm 0.012$ & $\textbf{0.940} \pm 0.017$ \\
Dslr & $0.961 \pm 0.027$ & $0.954 \pm 0.009$ & $\textbf{0.967} \pm 0.009$ \\
\\
\end{tabular}
\end{center}
\label{table:knn three runs}
\end{table}
\subsubsection{White-box Scenario} \label{A.6.5}
\paragraph{Office-31} 
We show the results with different mixing weight initialization in Table~\ref{table:office-31 more results}. $\lambda_{new}$ is initialized with $[0.01, 0.4, 0.5, 0.6, 0.99]$ for both $m=1$ and $m=2$. 
\begin{table}[ht]
\caption{\texttt{Office-31} in White-box scenario with different mixing weight initialization.}
\begin{center}
\small
\begin{tabular}{lcccc}
 Method & $A, D \rightarrow W$ & $A, W \rightarrow D$ & $W, D \rightarrow A$ & AVG\\
\midrule
Independent model & 69.61 & 75.62 & 71.73 & 72.34\\
Finetune Source & 91.25 & 88.24 & 82.69 & 87.39\\
[5pt]
m=1 $\lambda_{new}=0.01$ & 78.25 & 76.47 & 94.35 & 83.02\\
m=1 $\lambda_{new}=0.4$ & 96.25 & 90.20 & 92.58 & \textbf{93.01}\\
m=1 $\lambda_{new}=0.5$ & \textbf{97.50} & 88.24 & 91.52 & 92.42\\
m=1 $\lambda_{new}=0.6$ & 95.00 & 88.24 & 85.87 & 89.70\\
m=1 $\lambda_{new}=0.99$ & 92.50 & 94.12 & 68.20 & 84.94\\
[5pt]
m=2 $\lambda_{new}=0.01$ & 81.25  & 64.71 & \textbf{96.47} & 80.81\\
m=2 $\lambda_{new}=0.4$ & 93.75 & 92.16 & 90.81 & 92.24\\
m=2 $\lambda_{new}=0.5$ & 92.50 & \textbf{94.12} & 85.16 & 90.59\\
m=2 $\lambda_{new}=0.6$ & 96.26 & 92.16 & 83.75 & 90.72\\
m=2 $\lambda_{new}=0.99$ & 93.75 & 90.20 & 74.56 & 86.17\\
\end{tabular}
\end{center}
\label{table:office-31 more results}
\end{table}
\paragraph{CIFAR100} For a closer look at the results shown in Figures~\ref{fig:cifar100}, we also present the results in Tables~\ref{table:numofsources} and \ref{table:diff_init}, respectively.
\begin{table}[ht]
\caption{Influence of the number of source models on performance for \texttt{CIFAR100}.}
\begin{center}
\small
\begin{tabular}{lcccc}
 & 320 & 160 & 80 & 40\\
\midrule
Independent model & $63.44 \pm 0.54$ & $52.29  \pm 0.61$ & $50.79 \pm 0.79$ & $44.69 \pm 0.93$ \\
Finetune Source & $62.99 \pm 0.60$ & $55.81 \pm 0.54$ & $49.47 \pm 0.80$ & $41.68 \pm 0.99$\\
[5pt]
$m=1$ & $63.47 \pm 0.55$ & $\textbf{56.20} \pm 0.78$ & $51.56 \pm 0.72$ & $\textbf{47.01} \pm 0.99$\\
$m=3$ & $63.44 \pm 0.50$ & $55.85 \pm 0.63$ & $52.05 \pm 0.65$ & $46.19 \pm 0.94$\\
$m=5$ & $\textbf{63.83} \pm 0.43$ & $55.27 \pm 0.55$ & $\textbf{52.67} \pm 0.76$ & $46.49 \pm 1.43$\\
AVG & 63.58 & 55.77 & 52.09 & 46.56\\
\end{tabular}
\end{center}
\label{table:numofsources}
\end{table}
\begin{table}[ht]
\caption{Influence of different mixing weight initialization on \texttt{CIFAR100} when $m=5$.}
\begin{center}
\small
\begin{tabular}{lcccc}
 & 320 & 160 & 80 & 40\\
 \midrule
 Independent model & $63.44 \pm 0.54$ & $52.29 \pm 0.61$ & $50.79 \pm 0.79$ & $44.69 \pm 0.93$\\
[5pt]
$\lambda_{new} = 0.16$ & $63.83 \pm 0.43$ & $55.27 \pm 0.55$ & $\textbf{52.67} \pm 0.76$ &$\textbf{46.49} \pm 1.43$\\
$\lambda_{new} = 0.3$ & $63.84 \pm 0.56$ & $\textbf{56.43} \pm 0.72$ & $51.55 \pm 0.67$ & $45.52 \pm 1.22$\\
$\lambda_{new} = 0.4$ & $63.91 \pm 0.62$ & $53.28 \pm 0.66$ & $51.15 \pm 0.77$ & $45.35 \pm 0.99$\\
$\lambda_{new} = 0.6$ & $\textbf{64.20} \pm 0.58$ & $52.68 \pm 0.51$ & $50.98 \pm 0.81$ & $44.87 \pm 1.15$\\
AVG & $63.95$ & $54.42$ & $51.59$ & $45.58$ \\
\end{tabular}
\end{center}
\label{table:diff_init}
\end{table}
\subsubsection{Black-box Scenario} \label{A.6.6}
In Table~\ref{ctrl_balc_box}, we also provide results on \texttt{CTrL} under black-box setting for source models. A LeNet-5 model is used as target model. Random horizontal flip and vertical flip data augmentation is used to increase the size of training set to 150 and validation set to 90, a batch size of 128 is used during training in this setting.
\begin{table}[ht]
\caption{\texttt{CTrL} in black-box scenario.}
\centering
\small
\begin{tabular}{lccccc}
 Method & $m=1$ & $m=3$ & $m=5$ & API & AVG \\
\midrule
Independent & -& - & - & - & 45.14 \\
Finetune Source & -& - & - & - & 46.31 \\
Ours (w/o $\CL_{DC}$) & 47.12 & 47.23 & \textbf{47.42} & 46.40 & \textbf{47.04} \\
Ours (w/ $\CL_{DC}$) & \textbf{47.33} & \textbf{47.26} & 46.83 & \textbf{46.41} & 46.96\\ 
\end{tabular}
\label{ctrl_balc_box}
\end{table}
\subsubsection{Additional Results on OfficeHome Dataset} m\label{A.6.7}
The Office-Home dataset consists of images from 4 different domains: Artistic images, Clip Art, Product images and Real-World images. Each domain contains images of 65 object categories found typically in Office and Home settings. For each domain in OfficeHome, we randomly select $80\%$ of its data as training set, $10\%$ as validation set, and $10\%$ as test set. Data augmentation is performed by using random crops and random horizontal flips. All images have been rescaled to $256 \times 256$ and then cropped to $224 \times 224$. The results for OfficeHome can be found in Table \ref{officehome}. Our model achieves the highest performance in most cases.

\begin{table}[ht]
\caption{\texttt{OfficeHome} in white-box scenario.}
\centering
\small
\begin{tabular}{lccccc}
 Method & $A,C,R \rightarrow P$ & $A,C,P \rightarrow R$ & $C,R,P\rightarrow A$ & $A,R,P\rightarrow C$ & AVG \\
\midrule
Model stacking & 56.77 & 36.67  & 22.56  & 48.38  & 41.10  \\
Model soup & 15.48 & 12.80  & 10.63  & 14.91  & 13.46  \\
Multi-src SVM & 69.22 & 42.85 & 24.68 & 56.72 & 48.37  \\
MCW & 68.42 & 43.74  & 22.37  & 52.37  & 46.73  \\
DATE & 59.34 & 38.43  & 21.47  & 45.43  & 41.17  \\
DECISION & 53.87 & 37.21  & 21.86  & 42.51  & 38.86  \\
Independent & 71.69 & 46.22  & 25.82  & 57.21 & 50.24 \\
Finetune Source & 71.78 & 45.08  & 26.23 & \textbf{59.95}  & 50.76 \\
Ours (m=1) & 72.58  & 46.45  & 25.82  & 58.35  & 50.80  \\
Ours (m=3) & \textbf{73.03}  & \textbf{47.44}  & \textbf{29.51}  & 58.81  & \textbf{52.20} \\ 
\end{tabular}
\label{officehome}
\end{table}

\begin{table*}[p] 
\caption{The 40 Tasks created with \texttt{CIFAR100}.}
\begin{center}
\small
\begin{tabular}{llccc}
\hline\noalign{\smallskip}
Task ID & Classes\\
\noalign{\smallskip}
\hline
\noalign{\smallskip}
1 & [aquarium\_fish, flatfish, ray, shark, trout]\\
2 & [orchid, poppy, rose, sunflower, tulip]\\
3 & [beaver, dolphin, otter, seal, whale]\\
4 & [bottle, bowl, can, cup, plate]\\
5 & [bear, leopard, lion, tiger, wolf]\\
6 & [apple, mushroom, orange, pear, sweet\_pepper]\\
7 & [hamster, mouse, rabbit, shrew, squirrel]\\
8 & [baby, boy, girl, man, woman]\\
9 & [maple\_tree, oak\_tree, palm\_tree, pine\_tree, willow\_tree]\\
10 & [bicycle, bus, motorcycle, pickup\_truck, train]\\
11 & [clock, keyboard, lamp, telephone, television]\\
12 & [bed, chair, couch, table, wardrobe]\\
13 & [bee, beetle, butterfly, caterpillar, cockroach]\\
14 & [bridge, castle, house, road, skyscraper]\\
15 & [cloud, forest, mountain, plain, sea]\\
16 & [camel, cattle, chimpanzee, elephant, kangaroo]\\
17 & [fox, porcupine, possum, raccoon, skunk]\\
18 & [crab, lobster, snail, spider, worm]\\
19 & [crocodile, dinosaur, lizard, snake, turtle]\\
20 & [lawn\_mower, rocket, streetcar, tank, tractor]\\
21 & [hamster, cattle, maple\_tree, squirrel, chimpanzee]\\
22 & [house, bridge, bicycle, baby, lamp]\\
23 & [shark, oak\_tree, shrew, beaver, plate]\\
24 & [willow\_tree, crocodile, tiger, otter, telephone]\\
25 & [bus, aquarium\_fish, camel, skunk, apple]\\
26 & [bee, forest, tank, sweet\_pepper, fox]\\
27 & [snail, whale, clock, lion, cockroach]\\
28 & [rabbit, castle, pine\_tree, cloud, boy]\\
29 & [butterfly, road, rocket, skyscraper, wardrobe]\\
30 & [spider, crab, tractor, mouse, seal]\\
31 & [bed, elephant, beetle, keyboard, train]\\
32 & [plain, table, ray, worm, sea]\\
33 & [trout, bear, kangaroo, caterpillar, turtle]\\
34 & [motorcycle, bottle, orchid, chair, leopard]\\
35 & [dolphin, can, porcupine, cup, pickup\_truck]\\
36 & [poppy, wolf, pear, bowl, man]\\
37 & [snake, tulip, streetwar, palm\_tree, girl]\\
38 & [lawn\_mower, television, mushroom, lizard, raccoon]\\
39 & [orange, dinosaur, possum, lobster, flatfish]\\
40 & [mountain, couch, rose, woman, sunflower]\\
\hline
\end{tabular}
\end{center}
\label{table: cifar100 details}
\end{table*}
\setlength{\tabcolsep}{1.4pt}

\setlength{\tabcolsep}{4pt}
\begin{table*}[p]
\caption{Details of \texttt{CTrL} $S_{long}$ (tasks 1-50).}
\begin{center}
\small
\begin{tabular}{lllccc}
\hline\noalign{\smallskip}
Task ID & Dataset & Classes & \#Train & \# Val & \# Test\\
\noalign{\smallskip}
\hline
\noalign{\smallskip}
1 & mnsit & [3, 0, 5, 6, 8] & 5000 & 2500 & 4804 \\
2 & svhn & [9, 1, 3, 7, 0] & 25 & 15 & 5000\\
3 & svhn & [2, 9, 7, 1, 3] & 25 & 15 & 5000\\
4 & svhn & [5, 3, 0, 2, 9] & 5000 & 2500 & 5000\\
5 & fashion-mnist & [Sandal, Dress, T-shirt/top, Ankle boot, Pullover] & 25 & 15 & 5000\\
6 & fashion-mnist & [Ankle boot, Sneaker, Dress, Shirt, Trouser] & 25 & 15 & 5000\\
7 & svhn & [5, 4, 6, 3, 7] & 5000 & 2500 & 5000\\
8 & cifar100 & [man, squirrel, mouse, keyboard, maple\_tree] & 2250 & 250 & 500\\
9 & cifar10 & [horse, cat, bird, frog, dog] & 5000 & 2500 & 5000\\      
10 & fashion-mnist & [Coat, Ankle boot, Shirt, Pullover, Sneaker] & 5000 & 2500 & 5000\\
11 & mnist & [0, 7, 1, 9, 6] & 25 & 15 & 4938\\
12 & cifar10 & [bird, ship, airplane, dog, frog] & 5000 & 2500 & 5000\\
13 & cifar100 & [hamster, bear, dolphin, bicycle, road] & 25 & 15 & 500\\
14 & mnist & [9, 5, 0, 8, 3] & 5000 & 2500 & 4846\\
15 & fashion-mnist & [Sandal, Trouser, Shirt, Bag, Coat] & 5000 & 2500 & 5000\\
16 & cifar10 & [frog, bird, deer, automobile, horse] & 5000 & 2500 & 5000\\
17 & cifar10 & [deer, ship, truck, airplane, frog] & 5000 & 2500 & 5000\\
18 & svhn & [7, 0, 6, 5, 3] & 5000 & 2500 & 5000\\
19 & mnist & [8, 5, 7, 6, 4] & 25 & 15 & 4806\\
20 & mnist & [1, 4, 2, 0, 9] & 25 & 15 & 4962\\
21 & cifar100 & [whale, train, boy, crab, caterpillar] & 2250 & 250 & 500\\
22 & cifar100 & [rabbit, cattle, camel, cockroach, caterpillar] & 2250 & 250 & 500\\
23 & cifar100 & [orchid, road, spider, snake, caterpillar] & 25 & 15 & 500\\
24 & svhn & [3, 7, 8, 4, 5] & 25 & 15 & 5000\\
25 & cifar100 & [cloud, raccoon, baby, lamp, orange] & 2250 & 250 & 500\\
26 & cifar100 & [dolphin, castle, flatfish, house, whale] & 2250 & 250 & 500\\
27 & cifar100 & [skunk, chimpanzee, tractor, raccoon, lion] & 25 & 15 & 500\\
28 & svhn & [8, 6, 9, 2, 5] & 5000 & 2500 & 5000\\
29 & fashion-mnist & [Coat, Sandal, Bag, Ankle boot, Trouser] & 5000 & 2500 & 5000\\
30 & mnist & [9, 8, 2, 6, 7] & 5000 & 2500 & 4932\\
31 & cifar10 & [truck, ship, deer, bird, automobile] & 25 & 15 & 5000\\
32 & cifar10 & [airplane, horse, dog, bird, cat] & 25 & 15 & 5000\\
33 & svhn & [7, 8, 4, 5, 2] & 25 & 15 & 5000\\
34 & cifar100 & [castle, fox, shark, skunk, crocodile] & 2250 & 250 & 500\\
35 & cifar100 & [wardrobe, bridge, otter, lawn\_mower, telephone] & 25 & 15 & 500\\
36 & fashion-mnist & [Bag, Coat, T-shirt/top, Dress, Sandal] & 25 & 15 & 5000\\
37 & cifar10 & [deer, airplane, automobile, dog, cat] & 25 & 15 & 5000\\
38 & cifar10 & [cat, deer, frog, horse, truck] & 5000 & 2500 & 5000\\
39 & svhn & [0, 7, 8, 4, 3] & 25 & 15 & 5000\\
40 & mnist & [5, 7, 9, 2, 4] & 5000 & 2500 & 4874\\
41 & cifar100 & [wolf, dinosaur, ray, girl, crab] & 2250 & 250 & 500\\
42 & fashion-mnist & [Sandal, Coat, Sneaker, Trouser, Dress] & 25 & 15 & 5000\\
43 & cifar100 & [snail, rose, sweet\_pepper, worm, mushroom] & 25 & 15 & 500\\
44 & fashion-mnist & [Sneaker, Trouser, T\-shirt/top, Dress, Pullover] & 25 & 15 & 5000\\
45 & fashion-mnist & [Bag, Sandal, Shirt, Sneaker, Dress] & 25 & 15 & 5000\\
46 & mnist & [1, 4, 9, 6, 8] & 25 & 15 & 4914\\
47 & cifar10 & [frog, dog, truck, horse, bird] & 25 & 15 & 5000\\
48 & svhn & [0, 2, 4, 1, 7] & 5000 & 2500 & 5000\\
49 & cifar100 & [bed, tiger, lamp, road, caterpillar] & 2250 & 250 & 500\\
50 & cifar100 & [rocket, mouse, butterfly, road, cattle] & 2250 & 250 & 500\\
\hline
\end{tabular}
\end{center}
\label{table: ctrl slong 1}
\end{table*}
\setlength{\tabcolsep}{1.4pt}

\setlength{\tabcolsep}{4pt}
\begin{table*}[p] 
\caption{Details of \texttt{CTrL} $S_{long}$ (tasks 51-100).}
\begin{center}
\small
\begin{tabular}{lllccc}
\hline\noalign{\smallskip}
Task ID & Dataset & Classes & \#Train & \# Val & \# Test\\
\noalign{\smallskip}
\hline
\noalign{\smallskip}
51 & cifar100 & [trout, road, plain, television, wardrobe] & 25 & 15 & 500\\
52 & cifar100 & [ray, pear, lion, whale, wardrobe] & 25 & 15 & 500\\
53 & cifar10 & [bird, dog, deer, frog, truck] & 25 & 15 & 5000\\
54 & mnist & [0, 7, 4, 9, 6] & 25 & 15 & 4920\\
55 & fashion-mnist & [Shirt, Sneaker, Pullover, Trouser, Ankle boot] & 5000 & 2500 & 5000\\
56 & fashion-mnist & [Bag, Pullover, Sneaker, T\-shirt/top, Shirt] & 25 & 15 & 5000\\
57 & mnist & [7, 9, 3, 8, 0] & 25 & 15 & 4954\\
58 & cifar10 & [ship, airplane, frog, automobile, deer] & 25 & 15 & 5000\\
59 & mnist & [5, 6, 2, 4, 7] & 25 & 15 & 4832\\
60 & mnist & [5, 4, 2, 6, 0] & 25 & 15 & 4812\\
61 & svhn & [0, 2, 7, 3, 6] & 25 & 15 & 5000\\
62 & mnist & [9, 8, 7, 0, 4] & 25 & 15 & 4936\\
63 & cifar10 & [frog, horse, airplane, ship, automobile] & 25 & 15 & 5000\\
64 & mnist & [9, 1, 2, 8, 6] & 25 & 15 & 4932\\
65 & svhn & [7, 2, 5, 9, 1] & 25 & 15 & 5000\\
66 & cifar100 & [streetcar, beaver, sea, dolphin, butterfly] & 25 & 15 & 500\\
67 & svhn & [6, 3, 9, 7, 2] & 25 & 75 & 5000\\
68 & mnist & [9, 5, 7, 0, 6] & 25 & 15 & 4830\\
69 & fashion-mnist & [T\-shirt/top, Bag, Dress, Shirt, Sneaker] & 25 & 15 & 5000\\
70 & svhn & [8, 9, 4, 1, 6] & 25 & 15 & 5000\\
71 & svhn & [5, 1, 0, 7, 2] & 25 & 15 & 5000\\
72 & mnist & [5, 1, 4, 2, 7] & 25 & 15 & 4874\\
73 & cifar10 & [frog, cat, horse, truck, deer] & 25 & 15 & 5000\\
74 & mnist & [3, 9, 4, 2, 8] & 25 & 15 & 4956\\
75 & cifar10 & [ship, truck, deer, dog, bird] & 25 & 15 & 5000\\
76 & cifar100 & [pear, bus, fox, cloud, oak\_tree] & 25 & 15 & 500\\
77 & svhn & [1, 0, 2, 3, 7] & 25 & 15 & 5000\\
78 & svhn & [2, 8, 5, 4, 3] & 25 & 15 & 5000\\
79 & cifar100 & [rabbit, woman, girl, skyscraper, tuplip] & 25 & 15 & 500\\
80 & cifar100 & [trout, road, bridge, bowl, oak\_tree] & 25 & 15 & 500\\
81 & mnist & [5, 4, 8, 0, 6] & 25 & 15 & 4786\\
82 & fashion-mnist & [Pullover, Dress, Coat, Bag, T\-shirt/top] & 25 & 15 & 5000\\
83 & cifar10 & [bird, cat, deer, dog, automobile] & 25 & 15 & 5000\\
84 & cifar10 & [cat, dog, ship, frog, bird] & 25 & 15 & 5000\\
85 & cifar100 & [chimpanzee, camel, palm\_tree, leopard, spider] & 25 & 15 & 500\\
86 & mnist & [3, 4, 5, 9, 6] & 25 & 15 & 4832\\
87 & svhn & [8, 9, 4, 7, 2] & 25 & 15 & 5000\\
88 & cifar10 & [horse, cat, dog, airplane, automobile] & 25 & 15 & 5000\\
89 & mnist & [9, 8, 0, 1, 6] & 25 & 15 & 4912\\
90 & cifar100 & [possum, chair, bowl, mountain, cloud] & 25 & 15 & 500\\
91 & svhn & [7, 1, 5, 9, 6] & 25 & 15 & 5000\\
92 & cifar10 & [bird, cat, ship, frog, deer] & 25 & 15 & 5000\\
93 & mnist & [2, 8, 7, 5, 6] & 25 & 15 & 4824\\
94 & svhn & [6, 8, 2, 0, 5] & 25 & 15 & 5000\\
95 & fashion-mnist & [Sandal, Sneaker, Pullover, Ankle boot, Bag] & 25 & 15 & 5000\\
96 & svhn & [4, 9, 2, 0, 1] & 25 & 15 & 5000\\
97 & cifar100 & [beetle, motorcycle, skunk, bee, kangaroo] & 25 & 15 & 500\\
98 & cifar100 & [rabbit, butterfly, rose, pear, mushroom] & 25 & 15 & 500\\
99 & svhn & [8, 7, 4, 2, 3] & 25 & 15 & 5000\\
100 & fashion-mnist & [Trouser, Sneaker, Bag, T\-shirt/top, Pullover] & 25 & 15 & 5000\\
\hline
\end{tabular}
\end{center}
\label{table: ctrl slong 2}
\end{table*}
\setlength{\tabcolsep}{1.4pt}
\setlength{\tabcolsep}{8pt}
\begin{table}
\centering
\caption{The 6 tasks created with Tiny-ImageNet. Classes that exist in different tasks are color-coded to highlight the class sharing.}
\label{table:tinyImageNet Details}
\begin{tabular}{lc}
    & classes \\
    \hline
    
    task1 & \makecell{'n02124075', 'n04067472', 'n04540053', 'n04099969',                     'n07749582', \\
             'n01641577', 'n02802426', 'n09246464', 'n07920052', 'n03970156', \\
             'n03891332', 'n02106662', 'n03201208', 'n02279972', 'n02132136', \\
             'n04146614', 'n07873807', 'n02364673', 'n04507155', 'n03854065', \\
             'n03838899', 'n03733131', 'n01443537', 'n07875152', 'n03544143', \\
             \colorbox{blue!30}{'n09428293', 'n03085013', 'n02437312', 'n07614500', 'n03804744',} \\
             \colorbox{blue!30}{'n04265275', 'n02963159', 'n02486410', 'n01944390', 'n09256479',} \\
             \colorbox{blue!30}{'n02058221', 'n04275548', 'n02321529', 'n02769748', 'n02099712',} \\
             \colorbox{blue!30}{'n07695742', 'n02056570', 'n02281406', 'n01774750', 'n02509815',} \\
             \colorbox{blue!30}{'n03983396', 'n07753592', 'n04254777', 'n02233338', 'n04008634',}} \\ \hline
    task2 & \makecell{\colorbox{blue!30}{'n09428293', 'n03085013', 'n02437312', 'n07614500', 'n03804744',} \\
    \colorbox{blue!30}{'n04265275', 'n02963159', 'n02486410', 'n01944390', 'n09256479',}\\
    \colorbox{blue!30}{'n02058221', 'n04275548', 'n02321529', 'n02769748', 'n02099712',}\\
    \colorbox{blue!30}{'n07695742', 'n02056570', 'n02281406', 'n01774750', 'n02509815',}\\
    \colorbox{blue!30}{'n03983396', 'n07753592', 'n04254777', 'n02233338', 'n04008634',}\\
    \colorbox{Melon}{'n02823428', 'n02236044', 'n03393912', 'n07583066', 'n04074963',}\\
                 \colorbox{Melon}{'n01629819', 'n09332890', 'n02481823', 'n03902125', 'n03404251',}\\
                 \colorbox{Melon}{'n09193705', 'n03637318', 'n04456115', 'n02666196', 'n03796401',}\\
                 \colorbox{Melon}{'n02795169', 'n02123045', 'n01855672', 'n01882714', 'n02917067',}\\
                 \colorbox{Melon}{'n02988304', 'n04398044', 'n02843684', 'n02423022', 'n02669723'}}\\
                 \hline
    task3 & \makecell{\colorbox{Melon}{'n02823428', 'n02236044', 'n03393912', 'n07583066', 'n04074963',}\\
                 \colorbox{Melon}{'n01629819', 'n09332890', 'n02481823', 'n03902125', 'n03404251',}\\
                 \colorbox{Melon}{'n09193705', 'n03637318', 'n04456115', 'n02666196', 'n03796401',}\\
                 \colorbox{Melon}{'n02795169', 'n02123045', 'n01855672', 'n01882714', 'n02917067',}\\
                 \colorbox{Melon}{'n02988304', 'n04398044', 'n02843684', 'n02423022', 'n02669723',}\\
                 \colorbox{SpringGreen}{'n04465501', 'n02165456', 'n03770439', 'n02099601', 'n04486054',}\\
                 \colorbox{SpringGreen}{'n02950826', 'n03814639', 'n04259630', 'n03424325', 'n02948072',}\\
                 \colorbox{SpringGreen}{'n03179701', 'n03400231', 'n02206856', 'n03160309', 'n01984695',}\\
                 \colorbox{SpringGreen}{'n03977966', 'n03584254', 'n04023962', 'n02814860', 'n01910747',}\\
                 \colorbox{SpringGreen}{'n04596742', 'n03992509', 'n04133789', 'n03937543', 'n02927161'}} \\ \hline
    task4 & \makecell{\colorbox{SpringGreen}{'n04465501', 'n02165456', 'n03770439', 'n02099601', 'n04486054',}\\
                 \colorbox{SpringGreen}{'n02950826', 'n03814639', 'n04259630', 'n03424325', 'n02948072',}\\
                 \colorbox{SpringGreen}{'n03179701', 'n03400231', 'n02206856', 'n03160309', 'n01984695',}\\
                 \colorbox{SpringGreen}{'n03977966', 'n03584254', 'n04023962', 'n02814860', 'n01910747',}\\
                 \colorbox{SpringGreen}{'n04596742', 'n03992509', 'n04133789', 'n03937543', 'n02927161',}\\
                 \colorbox{Goldenrod}{'n01945685', 'n02395406', 'n02125311', 'n03126707', 'n04532106',}\\
                 \colorbox{Goldenrod}{'n02268443', 'n02977058', 'n07734744', 'n03599486', 'n04562935',}\\
                 \colorbox{Goldenrod}{'n03014705', 'n04251144', 'n04356056', 'n02190166', 'n03670208',}\\
                 \colorbox{Goldenrod}{'n02002724', 'n02074367', 'n04285008', 'n04560804', 'n04366367',}\\
                 \colorbox{Goldenrod}{'n02403003', 'n07615774', 'n04501370', 'n03026506', 'n02906734'}} \\ \hline
\end{tabular}
\end{table}
\setlength{\tabcolsep}{8pt}
\begin{table}
\centering
\caption{(Continued) The 6 tasks created with Tiny-ImageNet. Classes that exist in different tasks are color-coded to highlight the class sharing.}
\label{table:tinyImageNet Details Continue}
\begin{tabular}{lc}
    & classes \\
    \hline
    task5 & \makecell{\colorbox{Goldenrod}{'n01945685', 'n02395406', 'n02125311', 'n03126707', 'n04532106',}\\
                 \colorbox{Goldenrod}{'n02268443', 'n02977058', 'n07734744', 'n03599486', 'n04562935',}\\
                 \colorbox{Goldenrod}{'n03014705', 'n04251144', 'n04356056', 'n02190166', 'n03670208',}\\
                 \colorbox{Goldenrod}{'n02002724', 'n02074367', 'n04285008', 'n04560804', 'n04366367',}\\
                 \colorbox{Goldenrod}{'n02403003', 'n07615774', 'n04501370', 'n03026506', 'n02906734',}\\
                 'n01770393', 'n04597913', 'n03930313', 'n04118538', 'n04179913',\\
                 'n04311004', 'n02123394', 'n04070727', 'n02793495', 'n02730930',\\
                 'n02094433', 'n04371430', 'n04328186', 'n03649909', 'n04417672',\\
                 'n03388043', 'n01774384', 'n02837789', 'n07579787', 'n04399382',\\
                 'n02791270', 'n03089624', 'n02814533', 'n04149813', 'n07747607'}\\ \hline
    task6 & \makecell{'n03355925', 'n01983481', 'n04487081', 'n03250847', 'n03255030',\\
                 'n02892201', 'n02883205', 'n03100240', 'n02415577', 'n02480495',\\
                 'n01698640', 'n01784675', 'n04376876', 'n03444034', 'n01917289',\\
                 'n01950731', 'n03042490', 'n07711569', 'n04532670', 'n03763968',\\
                 'n07768694', 'n02999410', 'n03617480', 'n06596364', 'n01768244',\\
                 'n02410509', 'n03976657', 'n01742172', 'n03980874', 'n02808440',\\
                 'n02226429', 'n02231487', 'n02085620', 'n01644900', 'n02129165',\\
                 'n02699494', 'n03837869', 'n02815834', 'n07720875', 'n02788148',\\
                 'n02909870', 'n03706229', 'n07871810', 'n03447447', 'n02113799',\\
                 'n12267677', 'n03662601', 'n02841315', 'n07715103', 'n02504458'}
\end{tabular}
\end{table}
\end{document}

%% file: math_commands.tex
\usepackage{amsmath,amsfonts,bm}

\def\eqref#1{equation~\ref{#1}}

\def\1{\bm{1}}

\DeclareMathAlphabet{\mathsfit}{\encodingdefault}{\sfdefault}{m}{sl}
\SetMathAlphabet{\mathsfit}{bold}{\encodingdefault}{\sfdefault}{bx}{n}

\DeclareMathOperator*{\argmin}{arg\,min}

%% file: math_cmd.tex
\newcommand{\Av}{{\boldsymbol A}}

\newcommand{\Bv}{{\boldsymbol B}}

\newcommand{\cv}{{\boldsymbol c}}

\newcommand{\fv}{{\boldsymbol f}}

\newcommand{\hv}{{\boldsymbol h}}

\newcommand{\ov}{{\boldsymbol o}}

\newcommand{\xv}{{\boldsymbol x}}
\newcommand{\yv}{{\boldsymbol y}}

\newcommand{\thetav}{{\boldsymbol \theta}}

\newcommand{\lambdav}{{\boldsymbol \lambda}}

\newcommand{\phiv}{{\boldsymbol \phi}}

\newcommand{\CT}{\mathcal{T}}

\newcommand{\BR}{\mathbb{R}}

\newcommand{\CX}{\mathcal{X}}
\newcommand{\CY}{\mathcal{Y}}

\newcommand{\CL}{\mathcal{L}}

\newcommand{\beq}{\begin{equation}}
\newcommand{\eeq}{\end{equation}}
\newcommand{\beqs}{\begin{eqnarray}}
\newcommand{\eeqs}{\end{eqnarray}}
\newcommand{\barr}{\begin{array}}
\newcommand{\earr}{\end{array}}

%% file: cites.bib
@article{dohare2021continual,
  title={Continual backprop: Stochastic gradient descent with persistent randomness},
  author={Dohare, Shibhansh and Mahmood, A Rupam and Sutton, Richard S},
  journal={arXiv preprint arXiv:2108.06325},
  year={2021}
}

@article{Ash2019OnTD,
  title={On the Difficulty of Warm-Starting Neural Network Training},
  author={Jordan T. Ash and Ryan P. Adams},
  journal={ArXiv},
  year={2019},
  volume={abs/1910.08475}
}

@inproceedings{cong2020gan,
  title={Gan memory with no forgetting},
  author={Cong, Yulai and Zhao, Miaoyun and Li, Jianqiao and Wang, Sijia and Carin, Lawrence},
  booktitle={NeurIPS},
  volume={33},
  pages={16481--16494},
  year={2020}
}

@inproceedings{Perez2018FiLMVR,
  title={FiLM: Visual Reasoning with a General Conditioning Layer},
  author={Ethan Perez and Florian Strub and Harm de Vries and Vincent Dumoulin and Aaron C. Courville},
  booktitle={AAAI},
  year={2018}
}

@inproceedings{Liang2020DoWR,
  title={Do We Really Need to Access the Source Data? Source Hypothesis Transfer for Unsupervised Domain Adaptation},
  author={Jian Liang and D. Hu and Jiashi Feng},
  booktitle={International Conference on Machine Learning},
  year={2020}
}

@article{liang2021distill,
  title={Distill and fine-tune: Effective adaptation from a black-box source model},
  author={Liang, Jian and Hu, Dapeng and He, Ran and Feng, Jiashi},
  journal={arXiv preprint arXiv:2104.01539},
  volume={1},
  number={3},
  year={2021}
}

@inproceedings{ahmed2021unsupervised,
  title={Unsupervised multi-source domain adaptation without access to source data},
  author={Ahmed, Sk Miraj and Raychaudhuri, Dripta S and Paul, Sujoy and Oymak, Samet and Roy-Chowdhury, Amit K},
  booktitle={Proceedings of the IEEE/CVF Conference on Computer Vision and Pattern Recognition},
  pages={10103--10112},
  year={2021}
}

@inproceedings{NEURIPS2019_6048ff4e,
 author = {Lee, Joshua and Sattigeri, Prasanna and Wornell, Gregory},
 booktitle = {Advances in Neural Information Processing Systems},
 editor = {H. Wallach and H. Larochelle and A. Beygelzimer and F. d\textquotesingle Alch\'{e}-Buc and E. Fox and R. Garnett},
 pages = {},
 publisher = {Curran Associates, Inc.},
 title = {Learning New Tricks From Old Dogs: Multi-Source Transfer Learning From Pre-Trained Networks},
 volume = {32},
 year = {2019}
}

@inproceedings{
tong2021a,
title={A Mathematical Framework for Quantifying Transferability in Multi-source Transfer Learning},
author={Xinyi Tong and Xiangxiang Xu and Shao-Lun Huang and Lizhong Zheng},
booktitle={Advances in Neural Information Processing Systems},
editor={A. Beygelzimer and Y. Dauphin and P. Liang and J. Wortman Vaughan},
year={2021}
}

@article{fang2022source,
  title={Source-Free Unsupervised Domain Adaptation: A Survey},
  author={Fang, Yuqi and Yap, Pew-Thian and Lin, Weili and Zhu, Hongtu and Liu, Mingxia},
  journal={arXiv preprint arXiv:2301.00265},
  year={2022}
}

@article{veniat2020efficient,
  title={Efficient continual learning with modular networks and task-driven priors},
  author={Veniat, Tom and Denoyer, Ludovic and Ranzato, Marc'Aurelio},
  journal={arXiv preprint arXiv:2012.12631},
  year={2020}
}

@article{Verma2021EfficientFT,
  title={Efficient Feature Transformations for Discriminative and Generative Continual Learning},
  author={Vinay Kumar Verma and Kevin J Liang and Nikhil Mehta and Piyush Rai and Lawrence Carin},
  journal={2021 IEEE/CVF Conference on Computer Vision and Pattern Recognition (CVPR)},
  year={2021},
  pages={13860-13870},
}

@article{Kurmi2021DomainIA,
  title={Domain Impression: A Source Data Free Domain Adaptation Method},
  author={V. Kurmi and Venkatesh K. Subramanian and Vinay P. Namboodiri},
  journal={2021 IEEE Winter Conference on Applications of Computer Vision (WACV)},
  year={2021},
  pages={615-625}
}

@inproceedings{li2020model,
  title={Model adaptation: Unsupervised domain adaptation without source data},
  author={Li, Rui and Jiao, Qianfen and Cao, Wenming and Wong, Hau-San and Wu, Si},
  booktitle={Proceedings of the IEEE/CVF conference on computer vision and pattern recognition},
  pages={9641--9650},
  year={2020}
}

@article{Hou2021VisualizingAK,
  title={Visualizing Adapted Knowledge in Domain Transfer},
  author={Yunzhong Hou and Liang Zheng},
  journal={2021 IEEE/CVF Conference on Computer Vision and Pattern Recognition (CVPR)},
  year={2021},
  pages={13819-13828}
}

@article{Chen2021SelfSupervisedNL,
  title={Self-Supervised Noisy Label Learning for Source-Free Unsupervised Domain Adaptation},
  author={Weijie Chen and Luojun Lin and Shicai Yang and Di Xie and Shiliang Pu and Yueting Zhuang and Wenqi Ren},
  journal={2022 IEEE/RSJ International Conference on Intelligent Robots and Systems (IROS)},
  year={2021},
  pages={10185-10192}
}

@inproceedings{Liu2021GraphCB,
  title={Graph Consistency Based Mean-Teaching for Unsupervised Domain Adaptive Person Re-Identification},
  author={Xiaobing Liu and Shiliang Zhang},
  booktitle={International Joint Conference on Artificial Intelligence},
  year={2021}
}

@article{xiong2021source,
  title={Source data-free domain adaptation of object detector through domain-specific perturbation},
  author={Xiong, Lin and Ye, Mao and Zhang, Dan and Gan, Yan and Li, Xue and Zhu, Yingying},
  journal={International Journal of Intelligent Systems},
  volume={36},
  number={8},
  pages={3746--3766},
  year={2021},
  publisher={Wiley Online Library}
}

@inproceedings{Tarvainen2017MeanTA,
  title={Mean teachers are better role models: Weight-averaged consistency targets improve semi-supervised deep learning results},
  author={Antti Tarvainen and Harri Valpola},
  booktitle={NIPS},
  year={2017}
}

@inproceedings{Dong2021ConfidentAnchor,
 author = {Dong, Jiahua and Fang, Zhen and Liu, Anjin and Sun, Gan and Liu, Tongliang},
 booktitle = {Advances in Neural Information Processing Systems},
 editor = {M. Ranzato and A. Beygelzimer and Y. Dauphin and P.S. Liang and J. Wortman Vaughan},
 pages = {2848--2860},
 publisher = {Curran Associates, Inc.},
 title = {Confident Anchor-Induced Multi-Source Free Domain Adaptation},
 volume = {34},
 year = {2021}
}

@article{He2015DeepRL,
  title={Deep Residual Learning for Image Recognition},
  author={Kaiming He and X. Zhang and Shaoqing Ren and Jian Sun},
  journal={2016 IEEE Conference on Computer Vision and Pattern Recognition (CVPR)},
  year={2015},
  pages={770-778}
}

@article{Dosovitskiy2020AnII,
  title={An Image is Worth 16x16 Words: Transformers for Image Recognition at Scale},
  author={Alexey Dosovitskiy and Lucas Beyer and Alexander Kolesnikov and Dirk Weissenborn and Xiaohua Zhai and Thomas Unterthiner and Mostafa Dehghani and Matthias Minderer and Georg Heigold and Sylvain Gelly and Jakob Uszkoreit and Neil Houlsby},
  journal={ArXiv},
  year={2020},
  volume={abs/2010.11929}
}

@article{radford2019language,
  title={Language Models are Unsupervised Multitask Learners},
  author={Radford, Alec and Wu, Jeff and Child, Rewon and Luan, David and Amodei, Dario and Sutskever, Ilya},
  journal={ArXiv},
  year={2019}
}

@inproceedings{Houlsby2019ParameterEfficientTL,
  title={Parameter-Efficient Transfer Learning for NLP},
  author={Neil Houlsby and Andrei Giurgiu and Stanislaw Jastrzebski and Bruna Morrone and Quentin de Laroussilhe and Andrea Gesmundo and Mona Attariyan and Sylvain Gelly},
  booktitle={International Conference on Machine Learning},
  year={2019}
}

@article{li2021prefix,
  title={Prefix-tuning: Optimizing continuous prompts for generation},
  author={Li, Xiang Lisa and Liang, Percy},
  journal={arXiv preprint arXiv:2101.00190},
  year={2021}
}

@article{krizhevsky2009learning,
  author = {Krizhevsky, Alex},
  pages = {32--33},
  title = {Learning Multiple Layers of Features from Tiny Images},
  journal = {arXiv},
  year = 2009
}

@inproceedings{saenko2010adapting,
  title={Adapting visual category models to new domains},
  author={Saenko, Kate and Kulis, Brian and Fritz, Mario and Darrell, Trevor},
  booktitle={Computer Vision--ECCV 2010: 11th European Conference on Computer Vision, Heraklion, Crete, Greece, September 5-11, 2010, Proceedings, Part IV 11},
  pages={213--226},
  year={2010},
  organization={Springer}
}

@article{Li2020GradMixMT,
  title={GradMix: Multi-source Transfer across Domains and Tasks},
  author={Junnan Li and Ziwei Xu and Yongkang Wong and Qi Zhao and M. Kankanhalli},
  journal={2020 IEEE Winter Conference on Applications of Computer Vision (WACV)},
  year={2020},
  pages={3008-3016}
}

@article{Jin2021DeepTL,
  title={Deep Transfer Learning for Multi-source Entity Linkage via Domain Adaptation},
  author={Di Jin and Bunyamin Sisman and Hao Wei and Xin Dong and Danai Koutra},
  journal={Proc. VLDB Endow.},
  year={2021},
  volume={15},
  pages={465-477}
}

@article{Xu2018DeepCN,
  title={Deep Cocktail Network: Multi-source Unsupervised Domain Adaptation with Category Shift},
  author={Ruijia Xu and Ziliang Chen and Wangmeng Zuo and Junjie Yan and Liang Lin},
  journal={2018 IEEE/CVF Conference on Computer Vision and Pattern Recognition},
  year={2018},
  pages={3964-3973}
}

@article{Zhuang2019ACS,
  title={A Comprehensive Survey on Transfer Learning},
  author={Fuzhen Zhuang and Zhiyuan Qi and Keyu Duan and Dongbo Xi and Yongchun Zhu and Hengshu Zhu and Hui Xiong and Qing He},
  journal={Proceedings of the IEEE},
  year={2019},
  volume={109},
  pages={43-76}
}

@inproceedings{yao2010boosting,
  title={Boosting for transfer learning with multiple sources},
  author={Yao, Yi and Doretto, Gianfranco},
  booktitle={2010 IEEE computer society conference on computer vision and pattern recognition},
  pages={1855--1862},
  year={2010},
  organization={IEEE}
}

@inproceedings{sun2015subspace,
  title={Subspace distribution alignment for unsupervised domain adaptation.},
  author={Sun, Baochen and Saenko, Kate},
  booktitle={BMVC},
  volume={4},
  pages={24--1},
  year={2015}
}

@inproceedings{jiang2007instance,
    title = "Instance Weighting for Domain Adaptation in {NLP}",
    author = "Jiang, Jing  and
      Zhai, ChengXiang",
    booktitle = "Proceedings of the 45th Annual Meeting of the Association of Computational Linguistics",
    month = jun,
    year = "2007",
    address = "Prague, Czech Republic",
    publisher = "Association for Computational Linguistics",
    url = "https://aclanthology.org/P07-1034",
    pages = "264--271",
}

@article{pan2010domain,
  title={Domain adaptation via transfer component analysis},
  author={Pan, Sinno Jialin and Tsang, Ivor W and Kwok, James T and Yang, Qiang},
  journal={IEEE transactions on neural networks},
  volume={22},
  number={2},
  pages={199--210},
  year={2010},
  publisher={IEEE}
}

@inproceedings{herath2017learning,
  title={Learning an invariant hilbert space for domain adaptation},
  author={Herath, Samitha and Harandi, Mehrtash and Porikli, Fatih},
  booktitle={Proceedings of the IEEE conference on computer vision and pattern recognition},
  pages={3845--3854},
  year={2017}
}

@inproceedings{kulis2011you,
  title={What you saw is not what you get: Domain adaptation using asymmetric kernel transforms},
  author={Kulis, Brian and Saenko, Kate and Darrell, Trevor},
  booktitle={CVPR 2011},
  pages={1785--1792},
  year={2011},
  organization={IEEE}
}

@inproceedings{peng2019moment,
  title={Moment matching for multi-source domain adaptation},
  author={Peng, Xingchao and Bai, Qinxun and Xia, Xide and Huang, Zijun and Saenko, Kate and Wang, Bo},
  booktitle={Proceedings of the IEEE/CVF international conference on computer vision},
  pages={1406--1415},
  year={2019}
}

@inproceedings{venkateswara2017deep,
  title={Deep hashing network for unsupervised domain adaptation},
  author={Venkateswara, Hemanth and Eusebio, Jose and Chakraborty, Shayok and Panchanathan, Sethuraman},
  booktitle={Proceedings of the IEEE conference on computer vision and pattern recognition},
  pages={5018--5027},
  year={2017}
}

@article{knn,
  author={Cover, T. and Hart, P.},
  journal={IEEE Transactions on Information Theory}, 
  title={Nearest neighbor pattern classification}, 
  year={1967},
  volume={13},
  number={1},
  pages={21-27},
  doi={10.1109/TIT.1967.1053964}}

@article{netzer2011reading,
  title={Reading digits in natural images with unsupervised feature learning},
  author={Netzer, Yuval and Wang, Tao and Coates, Adam and Bissacco, Alessandro and Wu, Bo and Ng, Andrew Y},
  journal={NIPS},
  year={2011}
}

@article{xiao2017fashion,
  title={Fashion-mnist: a novel image dataset for benchmarking machine learning algorithms},
  author={Xiao, Han and Rasul, Kashif and Vollgraf, Roland},
  journal={arXiv preprint arXiv:1708.07747},
  year={2017}
}

@article{lecun2010mnist,
  title={MNIST handwritten digit database},
  author={LeCun, Yann and Cortes, Corinna and Burges, CJ},
  journal={ATT Labs [Online]. Available: http://yann.lecun.com/exdb/mnist},
  volume={2},
  year={2010}
}

@article{cortes1995support,
  title={Support-vector networks},
  author={Cortes, Corinna and Vapnik, Vladimir},
  journal={Machine learning},
  volume={20},
  pages={273--297},
  year={1995},
  publisher={Springer}
}

@article{Wortsman2022ModelSA,
  title={Model soups: averaging weights of multiple fine-tuned models improves accuracy without increasing inference time},
  author={Mitchell Wortsman and Gabriel Ilharco and Samir Yitzhak Gadre and Rebecca Roelofs and Raphael Gontijo-Lopes and Ari S. Morcos and Hongseok Namkoong and Ali Farhadi and Yair Carmon and Simon Kornblith and Ludwig Schmidt},
  journal={ArXiv},
  year={2022},
  volume={abs/2203.05482}
}

@article{Zhen2022OnTV,
  title={On the Versatile Uses of Partial Distance Correlation in Deep Learning},
  author={Xingjian Zhen and Zihang Meng and Rudrasis Chakraborty and Vikas Singh},
  journal={Computer vision - ECCV ... : ... European Conference on Computer Vision : proceedings. European Conference on Computer Vision},
  year={2022},
  volume={13686},
  pages={
          327-346
        }
}

@article{Szekely2007MeasuringAT,
  title={Measuring and testing dependence by correlation of distances},
  author={G'abor J. Sz'ekely and Maria L. Rizzo and Nail K. Bakirov},
  journal={Annals of Statistics},
  year={2007},
  volume={35},
  pages={2769-2794}
}

@article{hyvarinen2000independent,
  title={Independent component analysis: algorithms and applications},
  author={Hyv{\"a}rinen, Aapo and Oja, Erkki},
  journal={Neural networks},
  volume={13},
  number={4-5},
  pages={411--430},
  year={2000},
  publisher={Elsevier}
}

@article{hyvarinen1999fast,
  title={Fast and robust fixed-point algorithms for independent component analysis},
  author={Hyvarinen, Aapo},
  journal={IEEE transactions on Neural Networks},
  volume={10},
  number={3},
  pages={626--634},
  year={1999},
  publisher={IEEE}
}

@article{lecun1998gradient,
  title={Gradient-based learning applied to document recognition},
  author={LeCun, Yann and Bottou, L{\'e}on and Bengio, Yoshua and Haffner, Patrick},
  journal={Proceedings of the IEEE},
  volume={86},
  number={11},
  pages={2278--2324},
  year={1998},
  publisher={Ieee}
}

@article{He2021MaskedAA,
  title={Masked Autoencoders Are Scalable Vision Learners},
  author={Kaiming He and Xinlei Chen and Saining Xie and Yanghao Li and Piotr Doll'ar and Ross B. Girshick},
  journal={2022 IEEE/CVF Conference on Computer Vision and Pattern Recognition (CVPR)},
  year={2021},
  pages={15979-15988}
}

@inproceedings{Izmailov2018AveragingWL,
  title={Averaging Weights Leads to Wider Optima and Better Generalization},
  author={Pavel Izmailov and Dmitrii Podoprikhin and T. Garipov and Dmitry P. Vetrov and Andrew Gordon Wilson},
  booktitle={Conference on Uncertainty in Artificial Intelligence},
  year={2018}
}

@article{Matena2021MergingMW,
  title={Merging Models with Fisher-Weighted Averaging},
  author={Michael Matena and Colin Raffel},
  journal={ArXiv},
  year={2021},
  volume={abs/2111.09832}
}

@article{mcinnes2018umap,
  title={Umap: Uniform manifold approximation and projection for dimension reduction},
  author={McInnes, Leland and Healy, John and Melville, James},
  journal={arXiv preprint arXiv:1802.03426},
  year={2018}
}

@article{le2015tiny,
  title={Tiny imagenet visual recognition challenge},
  author={Le, Ya and Yang, Xuan},
  journal={CS 231N},
  volume={7},
  number={7},
  pages={3},
  year={2015}
}

@inproceedings{nguyen2020leep,
  title={Leep: A new measure to evaluate transferability of learned representations},
  author={Nguyen, Cuong and Hassner, Tal and Seeger, Matthias and Archambeau, Cedric},
  booktitle={International Conference on Machine Learning},
  pages={7294--7305},
  year={2020},
  organization={PMLR}
}

@inproceedings{you2021logme,
  title={Logme: Practical assessment of pre-trained models for transfer learning},
  author={You, Kaichao and Liu, Yong and Wang, Jianmin and Long, Mingsheng},
  booktitle={International Conference on Machine Learning},
  pages={12133--12143},
  year={2021},
  organization={PMLR}
}

@inproceedings{tran2019transferability,
  title={Transferability and hardness of supervised classification tasks},
  author={Tran, Anh T and Nguyen, Cuong V and Hassner, Tal},
  booktitle={Proceedings of the IEEE/CVF International Conference on Computer Vision},
  pages={1395--1405},
  year={2019}
}

@inproceedings{guo2023identifying,
  title={Identifying useful learnwares for heterogeneous label spaces},
  author={Guo, Lan-Zhe and Zhou, Zhi and Li, Yu-Feng and Zhou, Zhi-Hua},
  booktitle={International Conference on Machine Learning},
  pages={12122--12131},
  year={2023},
  organization={PMLR}
}

@article{deng2021flattening,
  title={Flattening sharpness for dynamic gradient projection memory benefits continual learning},
  author={Deng, Danruo and Chen, Guangyong and Hao, Jianye and Wang, Qiong and Heng, Pheng-Ann},
  journal={Advances in Neural Information Processing Systems},
  volume={34},
  pages={18710--18721},
  year={2021}
}

@inproceedings{Liebenwein2021SparseFP,
  title={Sparse Flows: Pruning Continuous-depth Models},
  author={Lucas Liebenwein and Ramin M. Hasani and Alexander Amini and Daniela Rus},
  booktitle={Neural Information Processing Systems},
  year={2021},
  url={https://api.semanticscholar.org/CorpusID:235623962}
}

@inproceedings{Cha2021SWADDG,
  title={SWAD: Domain Generalization by Seeking Flat Minima},
  author={Junbum Cha and Sanghyuk Chun and Kyungjae Lee and Han-Cheol Cho and Seunghyun Park and Yunsung Lee and Sungrae Park},
  booktitle={Neural Information Processing Systems},
  year={2021},
  url={https://api.semanticscholar.org/CorpusID:235367622}
}

@article{Li2017VisualizingTL,
  title={Visualizing the Loss Landscape of Neural Nets},
  author={Hao Li and Zheng Xu and Gavin Taylor and Tom Goldstein},
  journal={ArXiv},
  year={2017},
  volume={abs/1712.09913},
  url={https://api.semanticscholar.org/CorpusID:3693334}
}

@inproceedings{han2023discriminability,
  title={Discriminability and transferability estimation: a bayesian source importance estimation approach for multi-source-free domain adaptation},
  author={Han, Zhongyi and Zhang, Zhiyan and Wang, Fan and He, Rundong and Su, Wan and Xi, Xiaoming and Yin, Yilong},
  booktitle={Proceedings of the AAAI Conference on Artificial Intelligence},
  volume={37},
  pages={7811--7820},
  year={2023}
}

@article{Ahmed2021UnsupervisedMD,
  title={Unsupervised Multi-source Domain Adaptation Without Access to Source Data},
  author={Sk. Miraj Ahmed and Dripta S. Raychaudhuri and S. Paul and Samet Oymak and Amit K. Roy-Chowdhury},
  journal={2021 IEEE/CVF Conference on Computer Vision and Pattern Recognition (CVPR)},
  year={2021},
  pages={10098-10107},
  url={https://api.semanticscholar.org/CorpusID:233025033}
}

@article{Liang2021DINEDA,
  title={DINE: Domain Adaptation from Single and Multiple Black-box Predictors},
  author={Jian Liang and Dapeng Hu and Jiashi Feng and Ran He},
  journal={2022 IEEE/CVF Conference on Computer Vision and Pattern Recognition (CVPR)},
  year={2021},
  pages={7993-8003},
  url={https://api.semanticscholar.org/CorpusID:244800743}
}

@article{Yi2023WhenSD,
  title={When Source-Free Domain Adaptation Meets Learning with Noisy Labels},
  author={Li Yi and Gezheng Xu and Pengcheng Xu and Jiaqi Li and Ruizhi Pu and Charles Ling and Allan Mcleod and Boyu Wang},
  journal={ArXiv},
  year={2023},
  volume={abs/2301.13381},
  url={https://api.semanticscholar.org/CorpusID:256416103}
}

@phdthesis{li2022deep,
  title={Deep Neural Networks for Multi-Source Transfer Learning},
  author={Li, Keqiuyin},
  school={University of Technology Sydney},
  year={2022}
}

@article{fang2024source,
  title={Source-free unsupervised domain adaptation: A survey},
  author={Fang, Yuqi and Yap, Pew-Thian and Lin, Weili and Zhu, Hongtu and Liu, Mingxia},
  journal={Neural Networks},
  pages={106230},
  year={2024},
  publisher={Elsevier}
}

@inproceedings{wu2024h,
  title={H-ensemble: An Information Theoretic Approach to Reliable Few-Shot Multi-Source-Free Transfer},
  author={Wu, Yanru and Wang, Jianning and Wang, Weida and Li, Yang},
  booktitle={Proceedings of the AAAI Conference on Artificial Intelligence},
  volume={38},
  pages={15970--15978},
  year={2024}
}

@inproceedings{Rame2022ModelRR,
  title={Model Ratatouille: Recycling Diverse Models for Out-of-Distribution Generalization},
  author={Alexandre Ram'e and Kartik Ahuja and Jianyu Zhang and Matthieu Cord and L{\'e}on Bottou and David Lopez-Paz},
  booktitle={International Conference on Machine Learning},
  year={2022},
  url={https://api.semanticscholar.org/CorpusID:254877458}
}

@ARTICLE{9119820,
  author={Cui, Liangze and Zhang, Yao and Zhang, Runren and Liu, Qing Huo},
  journal={IEEE Transactions on Antennas and Propagation}, 
  title={A Modified Efficient KNN Method for Antenna Optimization and Design}, 
  year={2020},
  volume={68},
  number={10},
  pages={6858-6866},
  doi={10.1109/TAP.2020.3001743}}
